%% file: main.tex
\pdfoutput=1

\documentclass[11pt]{article}

\usepackage[]{acl}

\PassOptionsToPackage{numbers, compress}{natbib}

\usepackage{framed}
\usepackage{caption}
\usepackage{svg}
\usepackage{multirow}
\usepackage{multicol}
\usepackage{hyperref}
\usepackage{array}
\usepackage{longtable}
\usepackage{pythonhighlight}
\usepackage{color, colortbl}
\usepackage{xcolor}
\definecolor{LightCyan}{rgb}{0.8, 0.9, 1}
\usepackage{mathtools}

\usepackage[utf8]{inputenc} % allow utf-8 input
\usepackage[T1]{fontenc}    % use 8-bit T1 fonts
\usepackage{hyperref}       % hyperlinks
\usepackage{url}            % simple URL typesetting
\usepackage{booktabs}       % professional-quality tables
\usepackage{amsfonts}       % blackboard math symbols
\usepackage{nicefrac}       % compact symbols for 1/2, etc.
\usepackage{adjustbox}
\usepackage{multirow}
\usepackage{multicol}
\usepackage{makecell}
\usepackage{bm}
\usepackage{amsmath}
\usepackage{amsfonts}
\usepackage{bbm}
\usepackage{xspace}
\usepackage{subfigure}
\usepackage{arydshln}
\usepackage{enumitem}
\usepackage{xcolor}
\usepackage{pifont}
\usepackage[frozencache,cachedir=.]{minted}
\usepackage{algorithm, algpseudocode}
\usepackage{wrapfig}

\usepackage{times}
\usepackage{latexsym}
\usepackage{tcolorbox}

\usepackage[T1]{fontenc}
\usepackage[utf8]{inputenc}

\usepackage{microtype}
\usepackage{textcomp}
\definecolor{revision}{rgb}{0.0, 0.7, 0.3}
\def \name{\textsc{Fermi}\xspace}
\newcommand{\ms}[1]{\tiny{$\pm$#1}}
\definecolor{Gray2}{gray}{0.7}

\title{Few-shot Personalization of LLMs with Mis-aligned Responses}

\author{
        Jaehyung Kim$^{1}$\thanks{This work is done when Jaehyung was postdoc at CMU.} \quad
	Yiming Yang$^{2}$ \\
	 $^{1}$Yonsei University \quad 
  $^{2}$Carnegie Mellon University \\
	{\tt jaehyungk@yonsei.ac.kr} 
}

\begin{document}

\maketitle

\begin{abstract}
As the diversity of users increases, the capability of providing personalized responses by large language models (LLMs) has become increasingly important. 
Existing approaches have only limited successes in LLM personalization, due to the absence of personalized learning or the reliance on shared personal data.
This paper proposes a new approach for a \textbf{f}ew-shot p\textbf{er}sonalization of LLMs with their \textbf{mi}s-aligned responses (\name{}). 
Our key idea is to learn a set of personalized prompts for each user by progressively improving the prompts using LLMs, based on user profile (\textit{e.g.}, demographic information) and a few examples of previous opinions.
During an iterative process of prompt improvement, we incorporate the contexts of mis-aligned responses by LLMs, which are especially crucial for the effective personalization of LLMs.
In addition, we develop an effective inference method to further leverage the context of the test query and the personalized prompts.
Our experimental results demonstrate that \name{} significantly improves performance across various benchmarks, compared to best-performing baselines.\footnote{The code is available at \url{https://github.com/bbuing9/Fermi}.}
\end{abstract}
\input{intro}
\input{related}
\input{method}
\input{experiments}
\input{conclusion}

\bibliography{acl}

\clearpage
\input{appendix}

\end{document}

%% file: intro.tex
\section{Introduction}

The recent development of large language models (LLMs) has significantly accelerated progress in various NLP tasks, and yielded real-world applications used by millions of users, such as coding assistants and chatbots \citep{openai2022chatgpt, team2023gemini, touvron2023llama}.
As the use of LLMs by diverse users in real-world applications increases, \textit{personalization} of LLMs, \textit{i.e.}, steering LLMs' responses towards the unique needs or preferences of individual users becomes progressively important \citep{glaese2022improving, solaiman2021process}.
However, recent studies show that LLMs' responses are often biased toward certain groups but not suited for other diverse groups of users, and such biases cannot be fixed by providing simple instructions \citep{santurkar2023whose}.

Recent work in steering the responses of LLMs can be roughly divided into two categories. One category
is prompt engineering, which heuristically incorporates the user's information into the input prompts of LLMs \citep{salemi2023lamp, hwang2023aligning}.
The other category focuses on learning from other users' data \citep{li2023steerability, zhao2024group}. 
Both categories have their limitations: prompt engineering for each user would be too costly as it requires exploring the extensive search space of all possible prompts to find the personalized prompt.
Also, the learning-based category relies on the unrealistic assumption that personal data can be shared without violating privacy considerations. 

\input{figures/Figure1}

This paper addresses those limitations by introducing a new approach, namely \textbf{F}ew-shot P\textbf{er}sonalization of LLMs with \textbf{mi}s-aligned responses (\name{}).
Our high-level idea is to use LLM to progressively improve its input prompts based on a few examples of previous user opinions and profiles (\textit{e.g.}, demographics) in an iterative process. 
In addition to the current prompts' scores measured on given few-shot user opinions \citep{yang2024large}, \name{} incorporates the \textit{mis-aligned responses} (\textit{i.e.}, LLM’s responses with those prompts, which are inconsistent with given user opinions) as additional context. 
The contexts of mis-aligned responses include useful learning signals to update prompts such as the types of wrong predictions with the current prompts (see the empirical evidence in Section \ref{sec:4}).
Specifically, the iterative process of \name{} consists of three steps: (1) scoring the initial or current prompts with LLM, (2) updating the memory with high-scored prompts in the form of $<$prompt, score, context$>$ triplets, and (3) generating new improved prompts with LLM based on the updated memory.
In addition, we propose Retrieval-or-Prompt, a method to improve the inference on a given test query. 
Retrieval-or-Prompt selectively uses one of the personalized prompts obtained from the optimization, based on the context of the test query. 
An overview of \name{} is presented in Figure \ref{fig:method}.

We demonstrate the effectiveness of \name{} for few-shot personalization of LLMs, through extensive evaluations on various tasks including question-answering (QA), classification, regression, and generation.
For example, we observe that \name{} exhibited 6.8\% and 4.1\% average accuracy improvements on two multiple-choice QA datasets, constructed to evaluate the personalization of LLMs, compared to the previous state-of-the-art heuristic and optimization approaches, respectively.
We also found that the personalized prompts produced with one LLM are also effective on other LLMs, including both API-based and open-sourced ones, which is crucial for efficient deployment in practice. 
In addition, our in-depth analyses reveal why \name{} is more effective than other prompting methods and what are the important features of prompts for effective personalization of LLMs.
We hope our work provides useful insights for the research on LLM personalization, which becomes increasingly emerging and important for the future success of LLMs in real-world applications.

%% file: figures/Figure1.tex
\begin{figure*}[t]
\centering
\includegraphics[width=\textwidth]{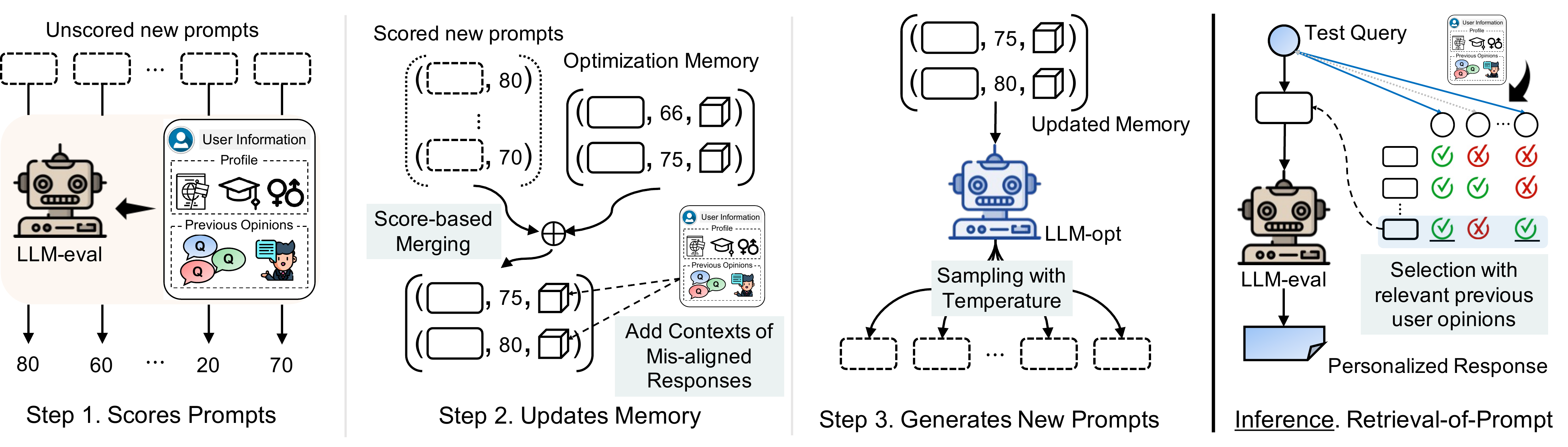}
\caption{
\textbf{An overview of \name{}.} \name{} iterates three steps to optimize the prompt from the given user information: (1) scoring new prompts, (2) updating the memory based on the score, and (3) generating new prompts (\textbf{left}). After the optimization, \name{} selectively uses the personalized prompts for the inference, via Retrieval-of-Prompt (\textbf{right}).
}
\label{fig:method}
\end{figure*}

%% file: related.tex
\section{Related Works}

\paragraph{Few-shot personalization of LLMs.}
Few-shot personalization of LLM is to align LLM's responses to a specific user with a limited number of user information such as user profile (\textit{e.g.}, demographic information) or opinions (\textit{e.g.}, previous responses to questions by user).
To this end, one line of prior works has explored how to input given user information into LLM in a heuristical manner, \textit{i.e.}, prompt engineering; for example, \citet{santurkar2023whose} designs three different templates of input prompt. 
\citet{salemi2023lamp} leverages the retrieval system \citep{izacard2022unsupervised} to use the given user opinions selectively. 
\citet{hwang2023aligning} shows that using both user profile and opinions is more effective.
While PEFT-based personalization method has been explored \citep{tan2024democratizing}, it’s hard to be applied to recent black-box LLMs.
On the other hand, another line of prior works has proposed learning from other user's data; 
\citet{li2023steerability} selects the relevant users using collaborative filtering, then learns the soft-prompt \citep{li2021prefix} from the augmented training data from these users' data. 
\citet{zhao2024group} proposes to train an independent transformer module via meta-learning on several users' data. 
However, both approaches have their limitations; prompt engineering incurs the cost of designing the prompt, and could be limited to fully utilizing the user information due to the absence of learning. 
The learning-based one necessitates other users' data which inevitably incurs privacy issues.
Therefore, we propose to only learn from target user's information and find the optimized (\textit{i.e.}, personalized) prompt for that user. 

\paragraph{Prompt optimization with LLM.}

As the prior works for prompt-tuning, relying on the gradient-based update \citep{deng2022rlprompt, lester2021power, shin2020autoprompt}, become inapplicable to the recent API-based LLM due to their black-box nature, other approaches have been recently explored for gradient-free prompt optimization, such as a progressive improvement using heuristic rules or LLMs \citep{prasad2023grips, yang2024large, zhou2023large}. 
For example, \citet{pryzant2023automatic} receives text feedback on how to update the prompts by instructing LLM. 
Also, after generating initial prompts with LLMs, \citet{zhou2023large} generates a semantically similar variant of the prompts with the highest accuracies.
\citet{yang2024large} iterates evaluation and generation of prompts with two LLMs, to solve the black-box optimization such as prompt optimization; \citet{yang2024large} incorporates the past generated prompts with their scores to enable the LLM for the optimization to construct new improved prompts.
However, only providing the scores on training examples is insufficient to optimize the prompt for few-shot personalization of LLMs, as the context with mis-aligned responses such as the types or patterns within recursively wrong predictions can't be captured in scores.
Therefore, we propose an efficient way to incorporate such context, along with an additional method to improve the inference by considering the context of the given query.

%% file: method.tex
\section{\name{}: Few-shot Personalization via Learning from Mis-aligned Responses}\label{sec3:method}

In this section, we present our framework proposed for \textbf{F}ew-shot P\textbf{er}sonalization of LLMs from \textbf{mi}s-aligned responses (\name{}).
We first present our problem setup in Section~\ref{sec3.1:pre}. 
Then, in Section~\ref{sec3.2:opt}, we present our core component that optimizes the input prompt with a given user information, by using LLM as a black-box optimizer along with the additional contexts from mis-aligned responses. 
Lastly, we introduce an efficient inference scheme after optimizing prompts with \name{}, by utilizing the context of a test query (Section~\ref{sec3.3:rop}). 

\subsection{Problem description}\label{sec3.1:pre}
We first describe the problem setup of our interest under a question-answering (QA) scenario.
Our goal is to steer LLM for a specific user using that user's information, and hence make LLM adaptively answer a given question depending on the user.
Formally, let ${q}$ denote the given test question and $\mathcal{M}$ denote the LLM, respectively. 
Next, for user $u$, we assume two types of user information: $U_{\tt pro}$ and $U_{\tt opi}$. 
$U_{\tt pro}$ indicates explicit profile of $u$ such as demographics information (\textit{e.g.}, region, sex, and age) or ideology (\textit{e.g.}, political affiliation). 
$U_{\tt opi}$ indicates $N$ {few-shot} previous opinions by $u$, which has the form of QA pairs, \textit{i.e.}, $U_{\tt opi}=\{({q}_{i},{a}_{i})\}_{i=1}^{N}$ where ${q}_{i}$ is a previously asked question and ${a}_{i}$ is an opinion (answer) by the user.
Then, for given test question $q$, our goal is to predict the answer $a$, which would be generated by user $u$, through LLM $\mathcal{M}$ using both $U_{\tt pro}$ and $U_{\tt opi}$. 
The heuristic design of input prompt $\text{p}$ to incorporate such user information has been previously explored \citep{hwang2023aligning, santurkar2023whose}, \textit{i.e.}, prediction $\widehat{a}$ is obtained by conditioning $\mathcal{M}$ with $\text{p}$ constructed using $U_{\tt pro}$ and $U_{\tt opi}$:
\begin{equation}\label{eq:pred}
    \widehat{a}(\text{p})=\mathcal{M}(q;\text{p}).
\end{equation}
However, heuristically designed prompts are limited to fully exploit the given user information; for example, the personalization of LLM was better with few user opinions (\textit{e.g.}, 3 or 8) compared to using all opinions in $u_{\tt opi}$ \cite{hwang2023aligning}.  
Therefore, we tackle this limitation by finding personalized prompts that align LLM to the user, through direct learning from given user information. 

\input{figures/Figure2}

\subsection{Effective prompt optimization with LLM from contexts of mis-aligned responses}\label{sec3.2:opt}
To mitigate the difficulties from the large scale and black-box nature of recent LLMs, we instead optimize input prompts to learn from user information. 
It is motivated by the recent work \citep{yang2024large} that uses two LLMs, $\mathcal{M}$ and $\mathcal{M}_{\tt opt}$, to solve black-box optimization, where $\mathcal{M}_{\tt opt}$ denotes another LLM used for the optimization.
Specifically, our key idea is incorporating the contexts of \textit{mis-aligned} responses (\textit{i.e.}, QAs in $U_{\tt opi}$ that $\mathcal{M}$ incorrectly predict with current prompts) during the optimization, instead of only using scores of the prompts (\textit{e.g.}, average accuracy of the prediction by $\mathcal{M}$ on $U_{\tt opi}$). 
As the contexts of mis-aligned responses include useful learning signals such as types or patterns of common wrong predictions, they could be effective in learning how to improve the prompts. 

\name{} starts with an initial prompt set $\text{P}^{0}=\{\text{p}^{0}\}$ and we adopt heuristically designed prompts in previous works \citep{hwang2023aligning, santurkar2023whose} for $\text{p}^{0}$.
Specifically, we use the user's explicit profile $U_{\tt pro}$ when it is available; thereby we fully utilize the given user information.
If not, we adopt vanilla prompt that instructs QA without user information (see details in Appendix \ref{app:ours}).

Then, \name{} iterate the following 3 steps: \textit{(1) Score Prompts}, \textit{(2) Update Memory}, and \textit{(3) Generate New Prompts}. 
See Figure \ref{fig:method} for the illustration.

\noindent$\circ$ \textit{Step 1: Score Prompts}. We first calculate the score ${s}_{k}$ of each prompt $\text{p}_{k} \in \text{P}^{t}$, by obtaining the predictions from $\mathcal{M}$ under $\text{p}_{k}$ and evaluating them using the user's previous answers:
\vspace{-0.01in}
\begin{equation}\label{eq:score}
    {s}_{k} = \sum\nolimits_{({q}_i,{a}_i) \sim U_{\tt opi}} \text{s}\big({a}_i, \widehat{a}_i(\text{p}_{k})\big) / N,
\end{equation}
where $\widehat{a}_i(\text{p}_{k})=\mathcal{M}({q}_i;\text{p}_{k})$.
Here, $\text{s}(\cdot,\cdot)$ is a specific metric to evaluate the prediction (\textit{e.g.}, accuracy). 
During this calculation of the score $s_k$ of the prompt $\text{p}_{k}$, we also collect 
pair of QA pairs $U^{k}_{\tt opi}$ that $\mathcal{M}$ yields \textit{mis-aligned response} $\widehat{a}_i$ to the user's answer $a_i$, under $\text{p}_{k}$:
\vspace{-0.01in}
\begin{equation}\label{eq:mis_aligned}
    U^{k}_{\tt opi} = \{({q}_i,{a}_i) \in U_{\tt opi}|\text{s}\big({a}_i, \widehat{a}_i(\text{p}_{k})\big)< \tau\},
\end{equation}
where $\tau$ is a threshold to judge the mis-alignment; for example, we set $\tau=0.5$ when we use the correctness of prediction as the score $\text{s}(\cdot,\cdot)$.

\noindent$\circ$ \textit{Step 2: Update Memory}. 
Next, we construct an optimization memory $M^{t}$, which is used for the input of $\mathcal{M}_{\tt opt}$ to generate new improved prompts, by providing the information of well-performing prompts through the contexts of their mis-aligned responses. 
To be specific, the optimization memory $M^{t}=\{(\text{p}_{l}, {s}_{l}, c_{l})\}_{l=1}^{L}$ is constructed by selecting top-$L$ prompts among $\text{P}^{t}$ and $M^{t-1}$ (where $M^{0}=\emptyset$), according to their scores (Eq. \ref{eq:score}).
Here, we present the triplets in $M^{t}$ in ascending order, \textit{i.e.}, ${s}_{l} < {s}_{l^{'}}$ when $l < l^{'}$, and provide the varied context $c_{l}$ depending on $l$.
Specifically, for $l=1$, we construct $c_{l}$ by concatenating QAs and mis-aligned responses by $\mathcal{M}$ under $\text{p}_{l}$ on $U^{l}_{\tt opi}$:
\vspace{-0.01in}
\begin{equation}\label{eq:context}
    c_{l} = \texttt{Concat}\{\big(i,{q}_{i},{a}_{i},\widehat{a}_{i}(\text{p}_{l})\big)|({q}_{i},{a}_{i})\in U^{l}_{\tt opi}\}.
\end{equation}

In Figure \ref{fig:prompt}, the texts corresponding to $c_1$ are highlighted in \textcolor{blue}{blue}. 
For other cases (\textit{i.e.}, $l\ne1$), instead of the enumeration like $c_{1}$, we construct the context $c_{l}$ with (i) the indices of \textit{common} mis-aligned QA pairs between $\text{p}_{l}$ and $\text{p}_{1}$, and (ii) the number of \textit{newly} mis-aligned QAs by $\text{p}_{l}$ compared to $\text{p}_{1}$  (see \textcolor{revision}{green texts} in Figure \ref{fig:prompt} for an example). 
Through the presented indices in $c_{l}$, $\mathcal{M}_{\tt opt}$ can directly access the mis-aligned QA pairs by referring $c_{1}$, and one can avoid unnecessary complexity of $c_{l}$ and cost from the long input to $\mathcal{M}_{\tt opt}$.
Additionally, the number of newly mis-aligned ones offers further insight into whether $\text{p}_{l}$ has improved, which can't be captured by the common mis-aligned ones.

\noindent$\circ$ \textit{Step 3: Generate New Prompts}. With the updated memory $M^{t}$, we generate $K$ new improved prompts $\text{P}^{t+1}=\{\text{p}^{\tt new}_{k}\}_{k=1}^{K}$ by prompting  $\mathcal{M}_{\tt opt}$ to generate the new and high-scored prompts:
\vspace{-0.01in}
\begin{equation}\label{eq:new_prompt}
    \text{p}^{\tt new}_{k} = \mathcal{M}_{\tt opt}(M^{t};\text{p}_{\tt opt}),
\end{equation}
where $\text{p}_{\tt opt}$ is a fixed input prompt for $\mathcal{M}_{\tt opt}$ to generate new prompts, and we use a random sampling with temperature to generate diverse new prompts from $\mathcal{M}_{\tt opt}$.  
Figure \ref{fig:prompt} presents the example of the overall input of $\mathcal{M}_{\tt opt}$ to generate new prompts, which is constructed with $M^{t}$ and $\text{p}_{\tt opt}$.

Then, we go back to Step 1 with $\text{P}^{t+1}$ and iterate these 3 steps for $T$ times. 
After that, we obtain the optimized (\textit{i.e.}, personalized) prompts $\text{P}^{T}=\{\text{p}^{T}_{k}\}_{k=1}^{K}$ for the user $u$. 

\subsection{Effective inference by Retrieval-of-Prompt}\label{sec3.3:rop} 
After $T$ iterations of the optimization procedure, \name{} outputs $K$ unique personalized prompts $\text{P}^{T}=\{\text{p}^{T}_{k}\}_{k=1}^{K}$.
Therefore, for a given test question $q$, one needs to determine which prompt to apply. 
Selecting the prompt with the highest score, \textit{i.e.}, $k^{*}=\arg\max_{k} s_{k}$ (Eq. \ref{eq:score}), would be a straight-forward way. 
However, our intuition is that better selection is possible if we utilize the context of the test question $q$ as additional information.
To this end, we propose to select the input prompt with the highest score on the subset of $U_{\tt opi}$, which only consists of the previous questions highly relevant to $q$.  
Formally, we first measure the relevance $r$ between $q$ and previous question ${q}_{i}$:
\vspace{-0.01in}
\begin{equation}\label{eq:relevance}
    R(q, U_{\tt opi})=\{r(q, {q}_{i})|{q}_{i} \in U_{\tt opi}\}.
\end{equation}
For the relevance $r$, we use the cosine similarity between the embeddings of questions, extracted by the sentence encoder \citep{reimers-2019-sentence-bert}. 
Then, we select top-$\tilde{N}$ questions according to the calculated relevance and construct the subset $U_{\tt opi}^{q}$ with those questions. 
Lastly, we choose the input prompt $\text{p}^{*}=\text{p}^{T}_{k^{*}}$ based on the score on $U_{\tt opi}^{q}$, which were already calculated, and use the prediction $\widehat{a}(\text{p}^{*})$ by $\mathcal{M}$:
\vspace{-0.01in}
\begin{equation}\label{eq:rop}
    k^{*}=\arg\max_{k} {s}^{T}_{k}( U_{\tt opi}^{q}),
\end{equation}
where ${s}^{T}_{k}(U_{\tt opi}^{q}) = \sum\nolimits_{({q}_i,{a}_i) \sim U_{\tt opi}^{q}} s\big({a}_i, \widehat{a}_i(\text{p}^{T}_{k})\big) / \tilde{N}$. 
Figure \ref{fig:method} illustrates the overview of \name{} and Algorithm \ref{alg:main} summarizes the overall procedure of \name{}.
We note that a full version of the prompts and examples of personalized prompts are presented in Appendixes \ref{app:setups} and \ref{app:more_examples}, respectively.

\input{our_algorithm}

%% file: figures/Figure2.tex
\begin{figure}[t]
\centering
\includegraphics[width=\columnwidth]{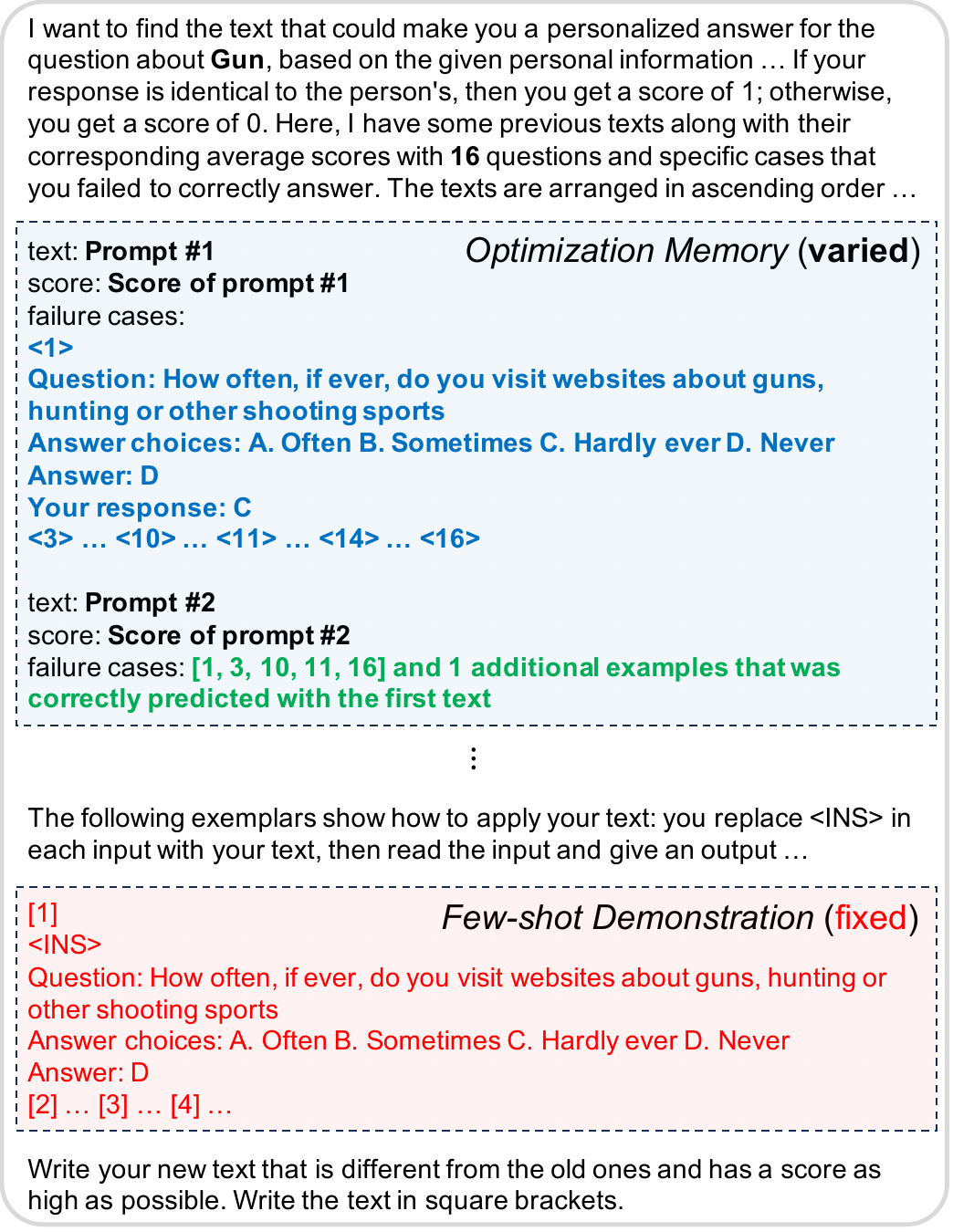}
%\vspace{-0.2in}
\caption{\textbf{Prompt example.} 
{Example of input prompt for $\mathcal{M}_{\tt opt}$ to generate new prompts, composed of fixed input prompt  $\text{p}_{\tt opt}$ (including fixed few-shot demonstrations) and optimization memory $M^{t}$ (Eq.~\ref{eq:new_prompt}) on OpinionQA dataset. A full version is in Appendix \ref{app:ours}.} 
}
%\vspace{-0.1in}
\label{fig:prompt}
\end{figure}

%% file: our_algorithm.tex
\begin{algorithm}[t!]
   \caption{\name{} algorithm}
   \label{alg:main}
\begin{algorithmic}
  \State
  \textbf{Input:} LLM for prediction $\mathcal{M}$, LLM for optimization $\mathcal{M}_{\tt opt}$, target test question $q$, explicit user profile $U_{\tt pro}$, previous user opinions $U_{\tt opi}=\{({q}_{i},{a}_{i})\}_{i=1}^{N}$, number of iterations $T$
  \vspace{0.01in} 
  \hrule
  \vspace{0.05in}  
  \State \texttt{\color{Gray2} /* Get initial prompt */}
  \State $\text{P}^{0}=\{\text{p}^{0}\}\leftarrow \texttt{InitPrompt}(U_{\tt pro})$ 
  \For{$t=0$ {\bfseries to} $T-1$}
  \State \texttt{\color{Gray2} /* Scoring prompts */} 
  \State $S^{t}=\{{s}_{k}\}_{k=1}^{K} \leftarrow \text{Eq.}~\ref{eq:score}$ with $\mathcal{M}$, $\text{P}^{t}$, $U_{\tt opi}$ 
  \State \texttt{\color{Gray2} /* Updating memory */} 
  \State $M^{t}=\{(\text{p}_{l}, {s}_{l}, c_{l})\}_{l=1}^{L} \leftarrow \text{Top-}L(M^{t-1}\cup\text{P}^{t}$) with $S^{t}$, $U_{\tt opi}$ (Eq. \ref{eq:context})
  \State \texttt{\color{Gray2} /* Generating new prompts */}
  \State $\text{P}^{t+1}=\{\text{p}^{\tt new}_{k}\}_{k=1}^{K} \leftarrow \text{Eq.~}\ref{eq:new_prompt}$ with $\mathcal{M}_{\tt opt}$, $M^{t}$ \hfill \texttt{\color{Gray2} /*Generate new prompts*/} 
  \EndFor
  \State \texttt{\color{Gray2} /* Retrieval-of-Prompt */} 
  \State $k^{*} \leftarrow \arg\max_{k} {s}^{T}_{k}(U_{\tt opi}^{q})$ (Eq.~\ref{eq:relevance})
  %\State $\widehat{a}(\text{p}^{*}) \leftarrow \mathcal{M}(q;\text{p}^{*}),~\text{p}^{*}=\text{p}^{T}_{k^{*}}$
  \State \textbf{return}~~$\widehat{a}(\text{p}^{*}) = \mathcal{M}(q;\text{p}^{T}_{k^{*}})$
\end{algorithmic}
\end{algorithm}

%% file: experiments.tex
\section{Experiments}\label{sec:4}

In this section, we design our experiments to investigate the following questions:
\begin{itemize}[leftmargin=5.5mm,topsep=0pt]
    \vspace{0.05in}
    \item[$\circ$] How does \name{} perform compare to other personalization methods? (Tables \ref{table:mcq} and \ref{table:lamp}) \vspace{-0.1in}
    \item[$\circ$] Is the optimized prompt with \name{} from one LLM transferable to different LLMs? (Table \ref{table:diff_llms}) \vspace{-0.28in}
    \item[$\circ$] What is the effect of each component in \name{}? (Table \ref{tab:ablation}) \vspace{-0.1in}
    \item[$\circ$] Why optimized prompt by \name{} is more effective than other prompts? (Table \ref{table:prompt_analyses})
\end{itemize}

\subsection{Setups}\label{sec:4.1}
First, we describe our experimental setups. More details are presented in Appendix \ref{app:setups}.

\noindent\textbf{Datasets.} For the experiments, we first use two multiple-choice QA datasets proposed to measure the steerability of LLMs for specific users (or social groups): \textit{OpinionQA} \citep{santurkar2023whose} and \textit{GlobalOpinionQA} \citep{durmus2023towards}.
For OpinionQA, we use a subsampled split released by \citet{hwang2023aligning}, which consists of 10.5k and 15.8k training and test QA pairs across 525 users and 15 topics, respectively.
For GlobalOpinionQA, since the dataset originally included the answer distribution by multiple respondents in the same country, we converted it to have a single answer by selecting the choice with the highest probability. 
It results in 920 training and 1,317 test QA pairs across 46 countries. 
We consider each country as a specific user. 
Next, we use three additional tasks, \textit{LaMP$_{\tt tag}$} (classification), \textit{LaMP$_{\tt rate}$} (regression), \textit{LaMP$_{\tt title}$} (generation), from a recent benchmark proposed for personalization of LLMs \citep{salemi2023lamp}. 
LaMP$_{\tt tag}$ is a 15-way classification task where an input is a movie description and a label is a movie tag, and LaMP$_{\tt rate}$ is a regression task where an input is a user review and a label is an integer rating (1-5).
LaMP$_{\tt title}$ is a generation task where the input is the abstract of the paper and the goal is generating the personalized title of the paper based on the given abstract.
We construct these datasets by subsampling from their original validation split, which results in 1,000 training and 1,500 test QA pairs across 50 users for each dataset. 
On average across four datasets, for each user, 20 training QAs as previous opinions and specific profile are given, and then 30 test QAs are used to evaluate.
Following \citet{salemi2023lamp}, we report average test accuracy (Acc), mean absolute error (MAE), and Rouge-L score for \textit{LaMP$_{\tt tag}$}, \textit{LaMP$_{\tt rate}$}, and \textit{LaMP$_{\tt title}$}, respectively.

\noindent\textbf{Baselines.} We compare \name{} against extensive baselines as follows: (1) \textit{Uniform}: expected performance when the prediction is made uniformly at random. 
(2) \textit{Vanilla}: answers the question with LLMs without any user information.
(3) \textit{Profile}: constructing prompt using all available user profiles \citep{santurkar2023whose, hwang2023aligning} such as demographics or nationality. 
(4) \textit{Few-shot}: retrieving relevant previous questions and opinions, then append them to the prompt \citep{hwang2023aligning, salemi2023lamp}. 
Following \cite{salemi2023lamp}, we consider BM25 \citep{robertson2009probabilistic} and Contriever \citep{izacard2022unsupervised} for the retriever models. 
The number of retrieved profiles is determined among \{3, 8, all\} with validation performance.
(5) \textit{All Info}: using both explicit profiles and retrieved previous QAs to construct prompt \citep{hwang2023aligning}. 
We use the retrieval with the best performance in \textit{Few-shot}.\footnote{In the case of OpinionQA, we additionally consider the retrieved indices originally included by \cite{hwang2023aligning}.}
(6) Optimization by PROmpting (\textit{OPRO}; \citet{yang2024large}): optimizing input prompt using both user profiles and previous opinions using LLMs.
Here, all of the previous opinions are utilized during the optimization. 
In the experiments, the prompt with the best training score is selected for the test.

\input{tables/Table1}
\input{tables/Table2}

\noindent\textbf{Implementation details.} 
We use three recent state-of-the-art LLMs for the prediction LLM $\mathcal{M}$ for the experiments: ChatGPT (\texttt{gpt-3.5-turbo-0613}) \citep{openai2022chatgpt}, GPT-4 (\texttt{gpt-4-turbo-1106}) \citep{openai2023gpt4}, Mistral-7B-Instruct-v0.2 \citep{jiang2023mistral} and LLaMA2-chat-70B \citep{touvron2023llama}. 
For $\mathcal{M}$, we use a temperature of $0.0$ when calling the API or greedy decoding for LLaMA, to remove the effect of random sampling.
For the optimization LLM $\mathcal{M}_{\tt opt}$, we always use GPT-4, as the prompt optimization based on the memory  (Eq.~\ref{eq:new_prompt}) requires complex reasoning capability (See Appendix \ref{app:more_analysis}), with a temperature of 1.0. 
For OPRO and \name{}, we use fixed values of $K=4$, $L=5$, and $T=10$. 
Also, with previous user opinions in $U_{\tt opi}$, 80\% is used for optimization and 20\% is used as few-shot demonstrations in $\text{p}_{\tt opt}$.
We set $\tau$, a threshold to define the mis-aligned responses (Eq. \ref{eq:score}), 0.2 only for \textit{LaMP$_{\tt title}$} and 1.0 for other tasks.
To obtain sentence embeddings for Retrieval-of-Prompt, we use the sentence encoder with MPNet \citep{song2020mpnet} showing the best performance.\footnote{Following the results in \url{https://www.sbert.net}} 
We use a fixed $\tilde{N}=3$ for Retrieval-of-Prompt.

\input{figures/Figure3}
\input{tables/Table3}

\subsection{Main results}

Table \ref{table:mcq} summarizes the experimental results on two different multiple-choice QA datasets, under ChatGPT.
First, it is observed that augmenting the user information into the input prompt is effective in improving the accuracies of LLMs, but the effectiveness could be varied.
For example, retrieving relevant user opinions is more effective than using the user profile for OpinionQA (49.8\% vs. 48.1\%), but it's vice versa in GlobalOpinionQA (61.2\% vs. 66.1\%).
It is due to the difference between datasets, as each user is asked multiple questions on the same topic in OpinionQA while GlobalOpinionQA asks the broader topics; 
this result also reveals the necessity of the learning-based prompt optimization approach.
From the results of OPRO and \name{}, one can observe that the optimization-based approach is actually effective, and the proposed method significantly improves it.
To be specific, \name{} exhibits 6.75\% average accuracy improvement compared to the previous prompting method. 
Furthermore, compared to the existing optimization method, \name{} exhibits 4.05\% accuracy improvement in the average.
In Figure \ref{fig:topic_acc}, we additionally present detailed results on OpinionQA, a topic-wise accuracy from four representative baselines selected based on average accuracy. 
Here, \name{} consistently shows better performance than other baselines across all topics, which further demonstrates the effectiveness of \name{} for the personalization of LLMs.

Next, Table \ref{table:lamp} summarizes the experimental results on LaMP$_{\tt tag}$, LaMP$_{\tt rate}$, and LaMP$_{\tt title}$. 
We note that these datasets do not include explicit user profiles; hence, we exclude both \textit{Profile} and \textit{All Info} for the baselines.
Here, it is noteworthy that the effectiveness of OPRO is significantly degraded, as the given task becomes more challenging to solve (\textit{e.g.}, the average number of answer choices: 3.96 for GlobalOpinionQA vs. 15 for LaMP$_{\tt tag}$). 
Nevertheless, \name{} is consistently effective and outperforms the other baselines; for example, \name{} exhibits 17.7\% and 4.0\% relative improvement in average, compared to the vanilla and the previous best baseline, respectively.

\subsection{Analyses with \name{}}\label{sec4.3}

In this section, we provide additional analyses of \name{} with the experiments on GlobalOpinionQA. 
More analyses are presented in Appendix \ref{app:more_analysis}. 

\paragraph{Transferability of the optimized prompt.}
Here, we provide additional experiments to verify the transferability of the learned prompt with our method.
To be specific, we first save the optimized prompts under ChatGPT as LLM for evaluation (Eq.~\ref{eq:pred}), which are used in Table \ref{table:mcq}.
Then, we directly apply these prompts to two different types of LLMs (Mistral-7B-Instruct-v0.2, LLaMA-2-chat-70B, and GPT-4), without additional optimization as same as applying heuristically designed prompts.
From Table \ref{table:diff_llms}, one can observe that the transferred prompts from \name{} significantly outperform the baseline prompting methods on both LLMs; for example, it exhibits 3.6\% accuracy improvement compared to the best-performing baseline on average. 
We remark that the prompts from OPRO are even less effective than the existing baseline, which further shows the advantages of \name{} in learning the well-generalized personalized prompt.  

\input{tables/Table4}
\input{tables/Table5}
\input{figures/Figure5}

\paragraph{Ablation study.} 
To validate the effectiveness of the proposed component of \name{} in Section \ref{sec3:method}, we perform the ablation experiments by decomposing our framework into three different components: (1) including QAs that have mis-alinged responses with the initial presentation and referring via common indices (\textit{Add$_{\tt Mis}$}), (2) noting the number of QAs with new mis-aligned responses (\textit{Add$_{\tt Num}$}), and (3) Retrieval-of-Prompt for a test query (\textit{RoP}). 
As shown in Table \ref{tab:ablation}, all components progressively improve the few-shot personalization of LLMs. 
Especially, it is observable that efficiently providing the context of mis-aligned QAs during the optimization is mostly crucial for the improvement.
Next, providing the number of new mis-aligned QAs makes additional improvement, as it can provide information about the effectiveness of the given prompt, which is not captured by commonly mis-aligned QAs. 
Lastly, for a test query, retrieving the most relevant prompt is more effective than selecting with the highest training score, as it successfully utilizes the context of the test query.

\paragraph{Features of good input prompts for personalization.} 
In Table \ref{table:prompt_analyses}, we further conduct the experiments to answer the following question: \textit{what features make good personalized prompts for LLMs?} 
First, we claim that the relevance of the prompt to the test query is crucial; for example, {Few-shot$_{\tt top3}$}, {Few-shot$_{\tt all}$}, and {Few-shot$_{\tt bott3}$} are different prompting methods by retrieving the 3 mostly relevant, all 20, and 3 mostly irrelevant previous opinions, respectively. 
Here, it is observable that test accuracy largely degrades when a portion of irrelevant opinions increases. 
Similarly, when we retrieved the most irrelevant prompt (\textit{\name{}$_{\tt irrel}$}), \textit{i.e.,} take $\arg\min$ in Eq.~\ref{eq:rop}, accuracy of \name{} is also decreased. 

Second, providing the user information with the proper format for LLMs is important. 
As shown in Figure \ref{fig:comparison_prompt}, the optimized prompt by \name{} is a detailed instruction consisting of multiple sentences that condense the lessons from the user opinions and LLM's mis-aligned responses. 
In contrast, the previous prompt used to incorporate previous opinions is based on the specific form, which is harder to follow by LLMs. 
To verify the importance of the format, we convert the enumeration of all QAs (by {Few-shot$_{\tt all}$}) into the instruction of multiple sentences (denoted by {Few-shot$_{\tt format}$}), by prompting GPT-4 using the optimized prompts by \name{} as reference. 
Interestingly, this format conversion shows significant improvement (56.3\% $\rightarrow$ 66.4\%) while it is still underperforming \name{}.

Lastly, effectively distilling the given user information is important.
As shown in Table \ref{table:prompt_analyses}, the prompting method with higher accuracy on previous user opinions $U_{\tt opi}$ (\textit{i.e.}, training accuracy) has a higher test accuracy for that user as well, except Few-shot$_{\tt all}$ which can directly access $U_{\tt opi}$.  
In this aspect, \name{} shows a clear advantage compared to the previous prompting optimization method; as shown in Figure \ref{fig:app_other_llms}, \name{} more effectively optimizes the prompt and achieves higher training accuracy than OPRO.
These results indicate that finding a proper way to condense and incorporate the user information to design input prompts is crucial, and \name{} achieves this by using the context of mis-aligned responses. 

Overall, designing personalized prompts satisfying these three properties {(relevancy to test query, proper format, and effective distillation of user information) is challenging, but \name{} effectively accomplishes this goal.

%% file: tables/Table1.tex
\begin{table}[t]
    \caption{\textbf{Main result on multiple-choice QA datasets.} Test accuracy of ChatGPT over the different methods on OpinionQA (OpQA) and GlobalOpinionQA (GOQA). The best and second best scores are highlighted in \textbf{bold} and \underline{underline}, respectively.}
    \vspace{-0.1in}
	\begin{center}
	\begin{adjustbox}{width=0.9\linewidth}
	\begin{tabular}{c|cc}
 		\toprule
            & \multicolumn{2}{c}{Datasets (Metric)} \\
		Methods & OpQA (Acc.) & GOQA (Acc.)   \\ \midrule
            Uniform & 34.2 & 31.4  \\  
		Vanilla & 45.5 & 62.8  \\ 
            Profile & 48.1 & 66.1 \\      
            Few-shot$_{\tt bm25}$ & 49.8 & 59.1 \\ 
            Few-shot$_{\tt cont}$ & 49.3 & 61.2 \\ 
            All Info & 48.6 & 62.3 \\ 
            OPRO & \underline{50.2} & \underline{71.1} \\ \midrule
            \name{} & \textbf{54.6} & \textbf{74.8} \\ \bottomrule
	\end{tabular}
    \end{adjustbox}
    \end{center}
    \label{table:mcq}
\end{table}

%% file: tables/Table2.tex
\begin{table}[t]
    \caption{\textbf{Main result on LaMP Benchmark.} Test performance of ChatGPT over the different methods on LaMP benchmarks. Test accuracy \big(Acc ($\uparrow$)\big), mean absolute error \big(MAE ($\downarrow$)\big), and Rouge-L ($\uparrow$) score are used, respectively. The best and second best scores are highlighted in \textbf{bold} and \underline{underline}, respectively.}
    \vspace{-0.1in}
	\begin{center}
	\begin{adjustbox}{width=0.9\linewidth}
	\begin{tabular}{c|ccc}
 		\toprule
   & \multicolumn{3}{c}{Datasets (Metric)} \\
		 & LaMP$_{\tt tag}$ & LaMP$_{\tt rate}$ & LaMP$_{\tt title}$   \\ 
  		Methods & (Acc.) & (MAE) & (Rouge-L)   \\ \midrule
  
    Uniform & 6.7 & 1.65 & - \\ 		
  Vanilla & 36.1 & 0.62 & 0.394\\ 
            Few-shot$_{\tt bm25}$ & {35.9} & {0.40} & \underline{0.411} \\  
            Few-shot$_{\tt cont}$ & \underline{36.2} & \underline{0.36} & 0.406 \\  
            OPRO & 34.3 & {0.57} & 0.406 \\ \midrule
            \name{} & \textbf{37.8} & \textbf{0.34} & \textbf{0.419}\\ \bottomrule
	\end{tabular}
    \end{adjustbox}
    \end{center}
    \label{table:lamp}
\end{table}

%% file: figures/Figure3.tex
\begin{figure*}[t]
\centering
%\vspace{-0.2in}
\includegraphics[width=1.0\textwidth]{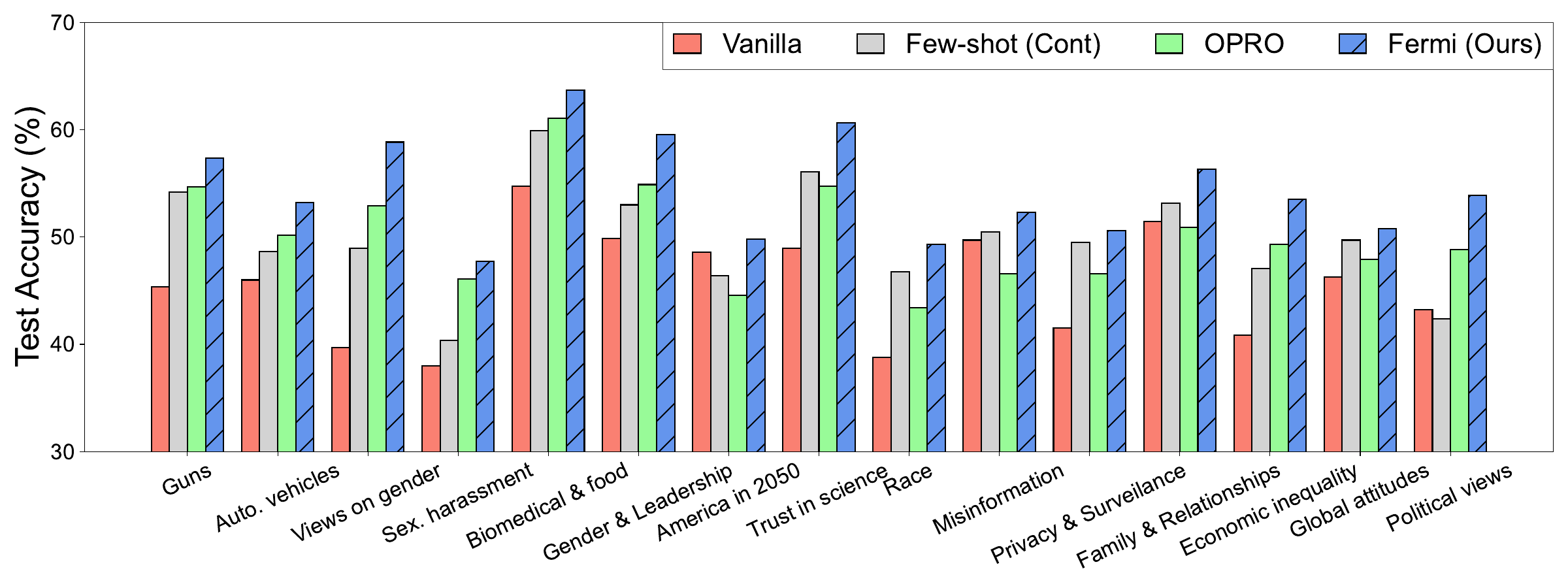}
\vspace{-0.1in}
\caption{\textbf{Overall topic-wise improvement.} Test accuracy of ChatGPT over four different personalization methods on OpinionQA. Detailed results are presented in Appendix \ref{app:more_results}.
}
\label{fig:topic_acc}
\end{figure*}

%% file: tables/Table3.tex
\begin{table}[t]
    \caption{\textbf{Transferability of \name{}.} Test accuracy of two  LLMs on GlobalOpinionQA. For Few-shot, we use Contriever which shows higher accuracy in Table \ref{table:mcq}. For OPRO$^{*}$ and \name{}$^{*}$, prompts optimized on ChatGPT are directly used. The best and second best scores are highlighted in \textbf{bold} and \underline{underline}, respectively.}
    \vspace{-0.1in}
	\begin{center}
	\begin{adjustbox}{width=0.9\linewidth}
	\begin{tabular}{c|ccc}
 		\toprule
		Methods & Mistral-v0.2 & LLaMA-2 & GPT-4 \\ \midrule
		Vanilla & 57.6 & 62.4 & 56.7 \\   
		Profile & 67.4 & \underline{65.5} & 77.7 \\   
		Few-shot & 70.1 & 60.5 & 68.9 \\   
            All Info & \underline{71.7} & 65.1 & \underline{78.2} \\
  		OPRO$^{*}$ & 66.4 & 64.5  & 76.7 \\ \midrule
            \name{}$^{*}$ & \textbf{72.4} & \textbf{68.9} & \textbf{84.8} \\ \bottomrule
	\end{tabular}
    \end{adjustbox}
    \end{center}
    \label{table:diff_llms}
\end{table}

%% file: tables/Table4.tex
\begin{table}[t]
\caption{\textbf{Ablation study of \name{}.} Test accuracy of ChatGPT on GlobalOpinionQA with different configurations of the proposed components in \name{}.}
\vspace{-0.1in}
        \begin{center}
	\begin{adjustbox}{width=0.9\linewidth}
%\scalebox{1.0}{
\begin{tabular}{c|ccc|cc}
\toprule
Methods & Add$_{\tt Mis}$ & Add$_{\tt Num}$ & RoP & Acc  \\ \midrule
OPRO                    & \textcolor{red}{\ding{55}}           &  \textcolor{red}{\ding{55}}                & \textcolor{red}{\ding{55}} & 71.1 \\
                 & \textcolor{green}{\ding{51}}           &  \textcolor{red}{\ding{55}}               & \textcolor{red}{\ding{55}} & 73.7 \\
                  & \textcolor{green}{\ding{51}}           &  \textcolor{green}{\ding{51}}                  & \textcolor{red}{\ding{55}}  & 74.2 \\ \midrule
\name{}                 & \textcolor{green}{\ding{51}}           &  \textcolor{green}{\ding{51}}                  & \textcolor{green}{\ding{51}}    & 74.8 \\ 
\bottomrule
\end{tabular}%}
\end{adjustbox}
\end{center}
\label{tab:ablation}
\end{table}

%% file: tables/Table5.tex
\begin{table}[t]
    \caption{\textbf{In-depth analyses about prompts for personalization.} Training and test accuracies of ChatGPT over the different methods on GlobalOpinionQA. Training accuracy is measured by given user opinions $u_{\tt opi}$.}
    \vspace{-0.1in}
	\begin{center}
	\begin{adjustbox}{width=0.8\linewidth}
	\begin{tabular}{c|cc}
 		\toprule
		Methods & Acc$_{\tt train}$ & Acc$_{\tt test}$ \\ \midrule
  Vanilla & 62.5 & 62.8 \\ 
            Profile & 67.9 & 66.1 \\ \midrule
            Few-shot$_{\tt top3}$ & - & 61.2 \\ 
            Few-shot$_{\tt all}$ & 95.2 & 56.3 \\ 
            Few-shot$_{\tt bott3}$ & - & 45.8 \\ 
            Few-shot$_{\tt format}$ & 70.2 & 66.4 \\ \midrule
            \name{}$_{\tt irrel}$ & 80.2 & 73.8 \\ 
            \name{} & 81.4 & {74.8} \\ \bottomrule
	\end{tabular}
    \end{adjustbox}
    \end{center}
    \label{table:prompt_analyses}
\end{table}

%% file: figures/Figure5.tex
\begin{figure*}[t]
\centering
\includegraphics[width=1.0\textwidth]{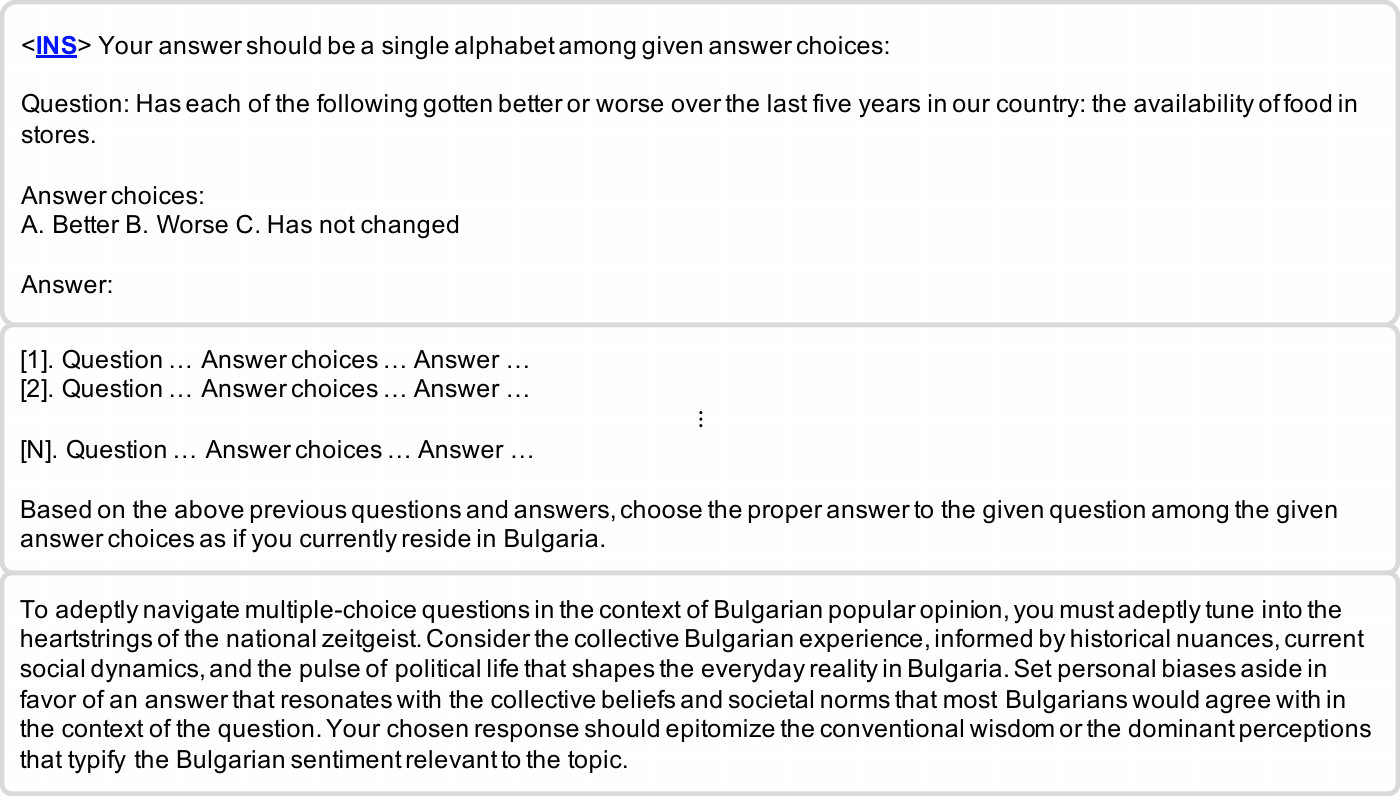}
\vspace{-0.1in}
\caption{
\textbf{Qualitative comparison.} Example prompts from All Info (\textbf{middle}) and \name{} (\textbf{bottom}) for the specific question (\textbf{top}) from GlobalOpinionQA. Prompt is inserted to <{\color{blue}\underline{\textbf{INS}}}>. More examples are in Appendix \ref{app:more_examples}.
}
\label{fig:comparison_prompt}
\end{figure*}

%% file: conclusion.tex
\section{Conclusion}

In this work, we propose \name{}, a simple yet effective framework for improving the few-shot personalization of LLMs.
Our key idea is to optimize the input prompt by learning from the user information; we propose an efficient way to incorporate contexts of mis-aligned responses by LLMs during the optimization, and a retrieval approach to select the optimized prompt relevant to test query. 
The effectiveness of \name{} is demonstrated by results on various personalization tasks and LLMs. 
\newpage
\section*{Limitations}

One possible limitation of \name{} is its computational cost.
As discussed in Appendix \ref{app:more_analysis}, our framework necessitates a strong LLM as the optimization of prompt requires complex reasoning capability. 
If we substitute $M_{\tt opt}$ from GPT-4 to ChatGPT, it can't properly optimize the input prompt; see Figure \ref{fig:app_other_llms} and we remark the similar observation was in the previous work \citep{yang2024large}.

Nevertheless, we would like to emphasize that the proposed \name{} is not just a simple consequence of more computations and costs. 
Compared to OPRO \citep{yang2024large}, another computationally intensive method for accuracy improvement, \name{} significantly outperforms it, \textit{i.e.}, \name{} is an even more efficient way to increase performance.
For instance, as shown in Figure \ref{fig:c}, \name{} achieves much higher accuracy than OPRO, (73.0\% v.s. 71.1\% on GlobalOpinionQA), even with half of the previous question and user’s responses and 1.4\% less cost per user (0.33\$ v.s. 0.34\$). 
The effectiveness of \name{} for optimizing personalized prompts is from more effective optimization by extracting the effective learning signal from mis-aligned responses. 
On the other hand, one can directly control the computational cost and performance, by varying the number of iterations ($T$) which linearly increases the cost with improved performance. 
Regarding this, we remark that 4 iterations of optimization with \name{} yield similar results to the 10 iterations of optimization with OPRO (see Figure \ref{fig:app_other_llms}). 

We further remark that the overall cost to use our framework will decrease while preserving its effectiveness, as the cost of using LLM with strong reasoning capability is continuously reduced. 
Currently, more than half of the overall cost occurs to use strong LLM (\textit{e.g.}, GPT4) for generating the improved prompts, and this choice is inevitable as this task requires the complex reasoning capability for LLM. 
However, as shown in Table \ref{table:app_mini}, we empirically observe that the recently released strong yet efficient LLM (GPT-4o-mini) can successfully optimize the prompt and yield a comparable performance with GPT-4, although it only requires 0.5\% input token and 1.0\% output per token price compared to GPT-4.   

At the same time, as we demonstrated in the experiments, the personalized prompts from our method are well-transferrable to other LLMs that are not used during optimization (Table \ref{table:diff_llms}), could be continuously updated with enlarged data through the user interactions (Table \ref{table:app_continual}), and also reusable to convert previous prompts to have the proper format for LLMs (Table \ref{table:prompt_analyses}).
Therefore, we believe that our approach could be an even more efficient way for personalization compared to the heuristical design of the prompt, after the consumption of the cost at the initial optimization.

\section*{Broader impact and ethical implications}

We strongly believe that \name{} can provide a strong positive impact in real-world applications that require personalized responses for the given user, \textit{e.g.}, search engines or chatbots.
We expect that our framework would be especially beneficial for the users belonging to under-populated social groups, since LLMs are known to follow the knowledge or opinion of the major population within pre-trained data \citep{kandpal2023large, santurkar2023whose}.
In contrast, there also exists some potential negative impacts. 
Since our framework needs to provide personal information to LLMs (mostly through API), it has a potential privacy risk when the provider of LLMs does not follow the safeguard and collects the given information.
In addition, as our framework didn't filter out the resulting prompts separately, it can include the prompts that have socially negative impacts, \textit{e.g.,} jailbreak of LLMs \citep{chao2023jailbreaking}. 
We believe that the incorporation of an additional filtering step could be a solution to this problem \citep{xie2023defending}.

%% file: appendix.tex
\appendix

\newpage
\section{More Analyses with \name{}}\label{app:more_analysis}

In this section, we provide more analyses of \name{} in addition to the analyses in Section \ref{sec4.3}.

\paragraph{GPT-4 for optimization $\mathcal{M}_{\tt opt}$.} 
As denoted in Section \ref{sec:4.1}, we commonly use GPT-4 for LLM $\mathcal{M}_{\tt opt}$ to generate new prompts from the optimization memory (Eq.~\ref{eq:new_prompt}) for all the experiments in Section \ref{sec:4}. 
To validate this design choice, we conduct the experiments by substituting GPT-4 with ChatGPT $\mathcal{M}_{\tt opt}$ in both OPRO and \name{}.
Figure \ref{fig:app_other_llms} is the optimization trajectory in terms of training accuracy (\textit{i.e.}, average accuracy of the prediction by $\mathcal{M}$ on previous user opinions).
Here, one can observe that both OPRO and \name{} suffer in optimizing the prompt when we use ChatGPT as $\mathcal{M}_{\tt opt}$, similar to the previous observation \cite{yang2024large}; it reveals that generating the improved prompts from the optimization memory with previous prompts, scores, and contexts requires complex reasoning capability.
Therefore, using a strong LLM with sufficient capability to optimize input prompts, such as GPT-4, is necessary.
But, we remark that one can substitute GPT-4 with cheaper LLMs with a sufficient reasoning capability (\textit{e.g.}, GPT-4o-mini), as shown in  Table \ref{table:app_mini}

\input{tables/App_Table_gpt4o_mini}

\input{figures/App_Figure_LLM_optim}
\input{tables/App_Table_GPT4}
\input{tables/App_Table_init_prompt}

\paragraph{Optimization with stronger LLM for evaluation  $\mathcal{M}$.} 
Next, to explore the compatibility of \name{} with different configurations of two LLMs during the optimization, we conduct the additional experiments by substituting evaluating LLM $\mathcal{M}$ to GPT-4 from ChatGPT; namely, two LLMs $\mathcal{M}$ and $\mathcal{M}_{\tt opt}$ for evaluating and generating are GPT-4. 
The results on GlobalOpinionQA are presented in Table \ref{table:app_gpt4}. 
It is observable that one can find further improved personalized prompts in terms of test accuracy, when using stronger LLM $\mathcal{M}$ for evaluating (Eq.~\ref{eq:score}).
For example, compared to the use of personalized prompts optimized by ChatGPT as $\mathcal{M}$ (\name{}$^{*}$), the optimization only using GPT-4 exhibits 1.9\% additional test accuracy improvement.
This result clearly shows that the proposed \name{} is compatible with different types and capacities of evaluating LLMs. 

\paragraph{Importance of initial prompts in \name{}.} 
For the experiments, we used a fixed initial prompt template across all datasets in our experiments, that maximally incorporates the given user profiles, as it has proven effective in prior studies \cite{hwang2023aligning, santurkar2023whose, zhao2024group}, as described in Section \ref{sec3.2:opt} and Appendix \ref{app:ours}.

Nevertheless, to further provide insights about the impact of initial prompt templates on \name{}, we conduct additional experiments by varying the initial prompt set $P^{0}=\{\text{p}_{0}\}$. 
To be specific, on GlobalOpinionQA dataset, we exclude the user profiles for the construction of the initial prompt unlike the original \name{} (in Table \ref{table:mcq}), and use the prompt of \textit{Vanilla} for the initialization. We denote this version as \name{}$_{\tt van}$.
The results (in comparison with other methods are) shown in Table \ref{table:app_no_init}, where \name{} (our method) consistently outperforms the baselines with both choices of prompt initialization while the gain is enlarged with better initialization when incorporating the user profile.  

\input{tables/App_Table_continual}
\input{figures/App_Figure_diff_hyper}

\paragraph{Continual optimization of prompts.} 
In the previous experiments, we assumed that the fixed dataset $U_{\tt opi}$ of questions and user's opinions is given. 
However, in the real-world, user often interacts frequently with LLMs, which means that the dataset could be continuously updated. 
Therefore, the iterative process of refining prompts might incur significant computational costs, if it should be conducted from scratch at certain intervals (\textit{e.g.}, when the number of new data reaches a threshold).

To mitigate this issue, we conduct additional experiments to show that the idea of continual prompt optimization \cite{razdaibiedina2023progressive, wang2022learning, zhu2022continual} could be applied to \name{}, and hence such cost could be drastically reduced. 
Specifically, we first conduct \name{} by using half of the previous questions and the user's responses $U_{\tt opi}$ (denoted by \name{}$^{\tt half}$). 
We remark that other parameters are kept the same such as the 10 iterations of the optimization. 
Then, with the entire $U_{\tt opi}$, we continuously conduct \name{} under the limited number of iterations, by initializing the prompt pool with previously optimized prompts in \name{}$^{\tt half}$ (\textit{i.e.}, substituting the initialization in line 116).
We denoted the results of this continuous optimization with 1 and 5 iterations as \name{}$^{\tt cont}_{\text{iter 1}}$ and \name{}$^{\tt cont}_{\text{iter 5}}$, respectively.

The results are presented in Table \ref{table:app_continual}. 
First, it is notable that even with the reduced number of data for the optimization, \name{} still outperforms the strong baselines that are based on heuristic prompt engineering (\textit{Profile}) or using the optimization by LLMs under full data (\textit{ORPO}). 
However, one can also observe that the accuracy under full data is much better (74.8 vs. 73.0), which reveals that the data quantity is still important in \name{}.
Next, it is also observed that the prompts could be successfully optimized continuously when the new data is added. 
Here, we denote that the previously optimized prompts in \name{}$^{\tt half}$ are also re-used for the pool of Retrieval-of-Prompt, to keep the knowledge of previous iterations.\footnote{Integrating new prompts into each user’s retrieval pool adds minimal computational overhead for calculating their embeddings.} 
Remarkably, even with only 1 additional iteration of optimization, the accuracy is significantly increased (73.0 $\rightarrow$ 74.0).
Also, when increasing the number of iterations to 5 (\textit{i.e,}, the same amount of computations compared to the original \name{}), the accuracy is increased and slightly outperforms the original optimization under the full data. 
Such improvement might be from an enlarged pool of Retrieval-of-prompt that enables better exploitation of previous knowledge.

These results clearly show that the proposed framework is still effective for a more realistic scenario under the continuously updated user data.

\noindent\textbf{Different hyper-parameter with \name{}.} 
Here, we first conduct the additional experiments by varying $K$ (number of new generation of prompts) and $L$ (number of prompts in the memory), and present these results in Figure \ref{fig:app_diff_hyper}.
Here, one can observe that using too small values of $K$ and $L$ significantly decreases the performance as it fails to find the effective prompt within the fixed iterations. 
However, after certain values including the originally used ones, \name{} successfully finds the personalized prompt and significantly outperforms the previous baselines. 

Next, we conduct new experiments that vary $N$ (number of previous questions and apply our framework, \name{}.
Here, one can observe that more previous questions continuously improve the personalization performance of our method. 
Nevertheless, one can also verify that \name{} is indeed more sample efficient; Fermi achieves much higher accuracy than OPRO (73.0\% v.s. 71.1\%), even using half of the user’s previous questions. 
This is because \name{} enables more efficient optimization by extracting the useful learning signal from mis-aligned responses, while OPRO just uses the average score of the current prompt as a learning signal.
Consequently, our framework is more sample-efficient than the previous state-of-the-art method (OPRO), and hence one can achieve better personalization results even with fewer previous questions. 

\iffalse
To address this issue, we believe that the idea of continual prompt optimization [4,5,6] could be applied to our framework. Specifically, when the amount of newly collected user opinions reaches a certain threshold, one can generate the updated prompts by initializing the prompt pool with previously optimized prompts (i.e., substituting the initialization in lines 226-228), instead of optimizing from scratch. Since the newly collected data originates from the same user who is considered for the previous optimization, a few iterations would be enough. It is also noteworthy that previous works have shown that such a continual learning framework is highly effective when the data distribution shift is relatively small [6,7].

We also like to emphasize that the proposed Retrieval-of-Prompt (presented in Section 3.3) could address the potential issue of forgetting previous knowledge during continual updates. By simply placing both previous and newly optimized prompts in the retrieval pool, the appropriate prompts can be automatically selected based on the context of the test query. Integrating new prompts into each user’s retrieval pool adds minimal computational overhead for calculating their embeddings.

We thank you once again for the constructive comments and will incorporate these discussions into the final version of the paper.
\fi

\section{Experimental Details}\label{app:setups}

\input{tables/App_Table1}

This section provides more details about the experimental setups in Section \ref{sec:4}.

\input{figures/App_Figure1}
\input{tables/App_Table_Global_List}

\subsection{Datasets}\label{app:datasets}

First, we present more detailed descriptions of the used datasets: OpinionQA \citep{santurkar2023whose}, GlobalOpinionQA \citep{durmus2023towards}, LaMP$_{\tt tag}$, LaMP$_{\tt rate}$, LaMP$_{\tt title}$ \citep{salemi2023lamp}. 
Dataset statistics are presented in Table \ref{table:app_dataset_statistics}.
Example of each dataset is presented in Figure \ref{fig:sup_dataset}.
\begin{itemize}[leftmargin=*,topsep=0.0pt,itemsep=.5pt]
    \item[$\circ$] \textbf{OpinionQA} is a multiple-choice QA dataset originally constructed based on a public opinion survey \citep{pewsurvey}, to evaluate the alignment of LM with 60 US demographic groups over various topics. 
    As OpinionQA includes the information of each respondent, this dataset has been also used to evaluate the personalization of LLMs \citep{hwang2023aligning} and we also adopt it. 
    Specifically, we use a subsampled split released by \citet{hwang2023aligning}, which consists of 10.5k and 15.8k training and test QA pairs across 525 users and 15 topics; namely, each user has 20 training QA pairs and 30 test QA pairs for each topic, on average.
    Also, the average number of answer choices is 3.2. 
    Then, we use training QA pairs as given previous opinions by user, and use test QA pairs to evaluate. In addition, for the experiments, we use all 12 types of user profiles included in the dataset: \{Age, Citizenship, Region, Education, Income, Marital status, Political ideology, Political party, Race, Religion, Frequency of religious attendance, Gender\}.
    \item[$\circ$] \textbf{GlobalOpinionQA} is a multiple-choice QA dataset constructed from cross-national surveys to capture diverse opinions on global issues across different countries. 
    Since the dataset originally included the answer distribution by multiple respondents in the same country, we converted it to have a single answer by selecting the choice with the highest probability, and treated each country as a specific user. 
    To be specific, we set a threshold (0.8) and selectively use the data when its highest probability is higher than the threshold to guarantee the quality of the converted. 
    It results in 920 training and 1,317 test QA pairs across 46 countries; namely, each user (country) has 20 training QA pairs and 28.6 test QA pairs for each topic, on average. 
    Also, the average number of answer choices is 4.1. 
    Then, we use training QA pairs as given previous opinions by user, and use test QA pairs to evaluate. 
    Also, nationality becomes the only available profile. 
    The full list of countries included in the dataset is presented in Table \ref{table:app_list_global}. Dataset could be downloaded from \url{https://huggingface.co/datasets/Anthropic/llm_global_opinions}.
    \item[$\circ$] \textbf{LaMP$_{\tt tag}$} is is a 15-way classification data where an input is a movie description and a label is a corresponding movie tag among 15 categories: \{Sci-fi, Based on a book, Comedy, Action, Twist ending, Dystopia, Dark comedy, Classic, Psychology, Fantasy, Romance, Thought-provoking, Social commentary, Violence, True story\}. 
    Since the original dataset is proposed to consider the scenario of fine-tuning LMs and hence it consists of a large number of examples, we construct our dataset by subsampling from its validation dataset to make it suitable to evaluate LLMs with inference. It results in 1,000 training and 1,500 test QA pairs across 50 users, respectively. 
    \item[$\circ$] \textbf{LaMP$_{\tt rate}$} is a regression data where an input is a user review and a label is an integer rating (1-5), \textit{i.e.}, 1 is mostly negative and 5 is mostly positive. Under the same motivation with LaMP$_{\tt tag}$, we construct our dataset by subsampling from its validation dataset, which results in 1,000 training and 1,500 test QA pairs across 50 users, respectively. 
    \item[$\circ$] \textbf{LaMP$_{\tt title}$} is a generation data where input is an abstract of the paper and a label is a title generated by the user. Under the same motivation with LaMP$_{\tt tag}$, we construct our dataset by subsampling from its validation dataset, which results in 1,000 training and 1,500 test QA pairs across 50 users, respectively. 
    LaMP benchmarks could be downloaded in \url{https://github.com/LaMP-Benchmark/LaMP}.
\end{itemize}

\subsection{Baselines}\label{app:baselines}

In this section, we present the specific prompts used for the experiments in Section \ref{sec:4}. 
Listing \ref{lst:base}-\ref{lst:both} are actually used prompts for Vanilla, Profile, Few-shot, and All Info, during the experiments on GlobalOpinionQA. 
Also, the prompt of OPRO used for the optimization is presented in Figure \ref{fig:app_opro_prompt}, which is the originally used one in \citet{yang2024large}. 
While we're trying to adapt this prompt similar to ours in Figure \ref{fig:app_full_prompt}, we observed that it degrades the performance of OPRO; for example, the average test accuracy is reduced to 70.7\% from 71.1\%. 
Therefore, we use the original prompt for all the experiments.
We remark that each prompt is minimally adjusted to consider the difference between datasets. 
For example, as OpinionQA includes many available user profiles, we fully incorporate these with the prompt in Listing \ref{lst:explicit_opqa}, following \citet{hwang2023aligning}. 
Also, we present the prompt of Vanilla method on LaMP$_{\tt rate}$ dataset in Listing \ref{lst:base_rate}. 
In addition, we present the prompt used to convert the format of input prompt by Few-shot (Table \ref{table:prompt_analyses}) in Listing \ref{lst:conversion}.

\subsection{\name{}}\label{app:ours}
\input{figures/App_Figure_full_prompt}

As denoted in Section \ref{sec3.2:opt}, we need to provide an initial input prompt set $\text{P}^{0}=\{\text{p}^{0}\}$. 
\iffalse
To this end, we use the heuristically design input prompt to incorporate user profiles (\textit{Profile} in Section \ref{sec:4}). 
Since our framework only utilizes a given few-shot previous opinions during the optimization, this way of initial prompting naturally enables us to fully utilize all the user information. 
To be specific, we adopt the prompts used for \textit{Profile} tuned for each data, which are presented in \ref{app:baselines}.
\fi
To this end, we use the heuristically design input prompts, which are presented in \ref{app:baselines}.
Specifically, we adopt the prompts used for \textit{Profile} tuned for each data, when the user profile $U_{\tt pro}$ is available (both OpinionQA and GlobalOpinionQA).
Since our framework only utilizes a given few-shot previous opinions during the optimization, this way of initial prompting naturally enables us to fully utilize all the user information.
When the user profile is not available, we adopt the prompts used for \textit{Vaniall}. 
In addition, we present a more detailed version of the prompt $\text{p}_{\tt opt}$ used to generate new input prompts with $\mathcal{M}_{\tt opt}$ in Figure \ref{fig:app_full_prompt}. 
We remark that $\text{p}_{\tt opt}$ is minimally adjusted across dataset, to match the different task and user information of each dataset.

\input{prompts/Vanilla}
\input{prompts/Explicit_global}
\input{prompts/Implicit}
\input{prompts/Both}

\input{figures/App_Figure_opro_prompt}

\input{prompts/Explicit_opqa}
\input{prompts/Vanilla_rate}

\input{prompts/Conversion}

\section{Additional Quantitative Results}\label{app:more_results}

In this section, we provide additional quantitative results that can't be presented in the main draft due to the limited space. 
First, in Table \ref{table:app_topic_acc}, we present the average and standard deviation of topic-wise accuracy, \textit{i.e.}, the average and standard deviation are calculated across 35 users where each user receives 30 test questions in the same topic.
Next, we present the test performance of \textit{Few-shot} method in Section \ref{sec:4}, under different numbers of retrieved opinions (Table \ref{table:app_diff_retrieve}).
Lastly, we present the test performance under a different number of considered training questions $\tilde{N}$ (Eq.~\ref{eq:rop}). 
As one can see in Table \ref{table:app_diff_rop}, $\tilde{N}=3$ which is commonly used in our experiments shows consistent improvements in general, although the optimal values are different across the datasets.

\input{tables/App_Table_Topics}

\input{tables/App_Table_Retrieve}
\input{tables/App_Table_rop}

\input{tables/App_lamp5}

\section{More Comparison Examples between Personalized Prompts}\label{app:more_examples}

In this section, we present more qualitative comparisons between the prompts from different methods for personalization of LLMs. 
To be specific, we present the specific test query from each data, and three corresponding prompts from the heuristic design, OPRO, and \name{}.
Figures \ref{fig:sup_example_op1}-\ref{fig:sup_example_rate2} are the comparison results on the datasets used in Section \ref{sec:4}.
Somewhat interestingly, one can observe that the personalized prompts by \name{} exhibit non-trivial incorporation of user information. 
In addition, we present examples of format-converted versions of few-shot prompting of previous user opinions (\textit{i.e.}, Few-shot$_{\tt format}$ in Table \ref{table:prompt_analyses}) in Figures \ref{fig:sup_example_conversion1} and \ref{fig:sup_example_conversion2}.
Here, one can observe that the converted prompts have a similar form to the personalized prompts by \name{} which is more natural to understand and follow for LLMs, and hence it significantly improves the performance up to 10.1\%, as shown in Table \ref{table:prompt_analyses}.
Figure \ref{fig:app_lamp5} is the example of personalized prompt and generated response under this on LaMP$_{\tt title}$.

\input{figures/App_Figure_examples_opinion}
\input{figures/App_Figure_examples_global}
\input{figures/App_Figure_examples_lamp_tag}
\input{figures/App_Figure_examples_lamp_rate}
\input{figures/App_Figure_examples_conversion}

%% file: tables/App_Table_gpt4o_mini.tex
\begin{table}[t]
%    \vspace{-0.2in}
    \caption{\textbf{Different LLM for $\mathcal{M}_{\tt opt}$.} Test accuracy of ChatGPT over the baseline methods and \name{} with the different LLMs (GPT-4o-mini and GPT-4) for generating prompts on GlobalOpinionQA.}
    \vspace{-0.1in}
	\begin{center}
	\begin{adjustbox}{width=1.0\linewidth} % Adjusted width
	\begin{tabular}{c|cc|cc}
 		\toprule
            & \multicolumn{4}{c}{Methods} \\
            Models & Vanilla & Profile &  \name{}$_{\texttt{mini}}$ & \name{}\\ \midrule
            ChatGPT & 62.8 & 66.1 & 74.2 & {74.8} \\ \bottomrule
	\end{tabular}
    \end{adjustbox}
    \end{center}
%    \vspace{-0.1in}
    \label{table:app_mini}
\end{table}

%% file: figures/App_Figure_LLM_optim.tex
\begin{figure}[t]
\centering
\includegraphics[width=1.0\columnwidth]{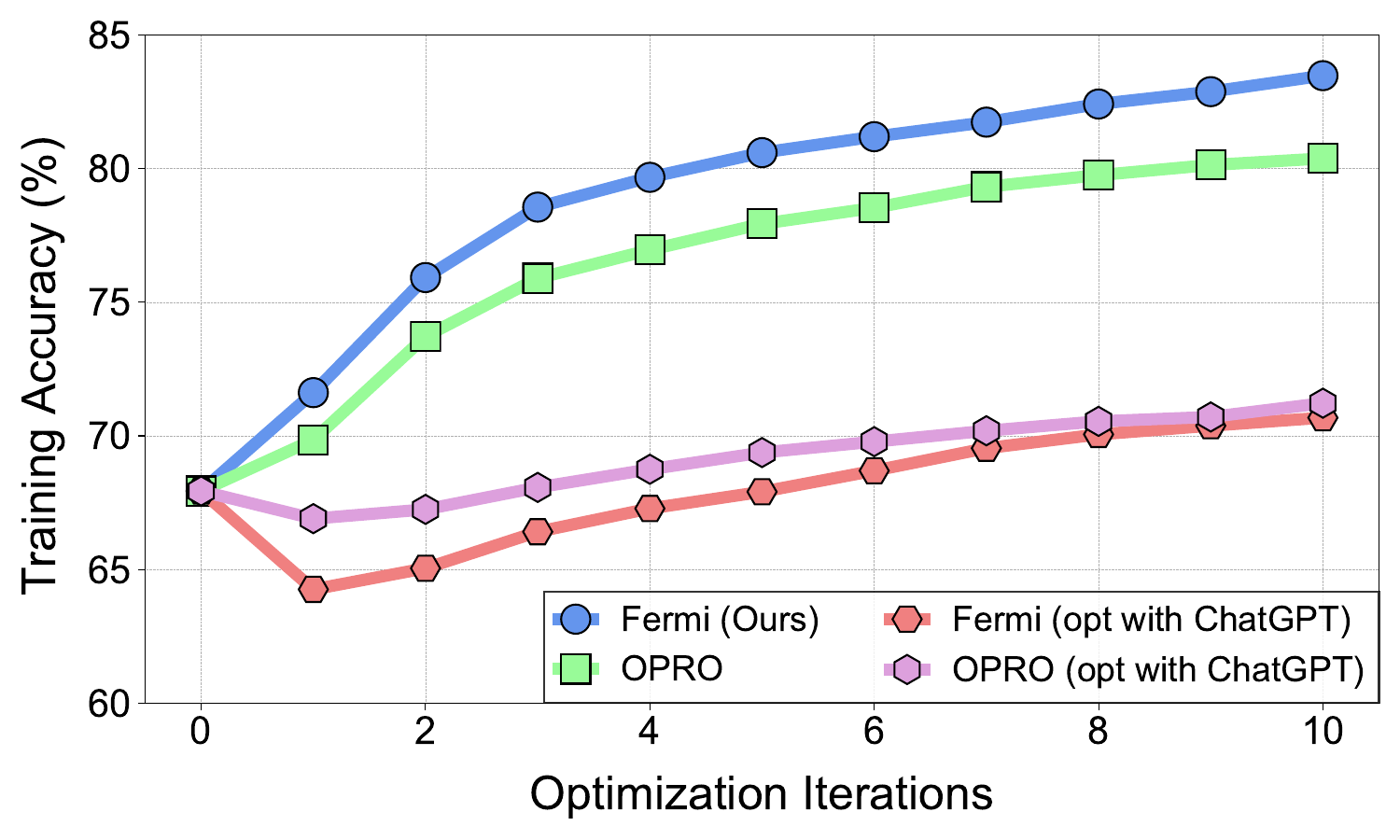}
%\vspace{-0.2in}
\vspace{-0.2in}
\caption{\textbf{Optimization trajectory under different LLMs for $\mathcal{M}_{\tt opt}$.} Average training accuracies on GlobalOpinionQA across optimization iterations ($T=10$) under OPRO and \name{}.
}
\vspace{-0.1in}
\label{fig:app_other_llms}
\end{figure}

%% file: tables/App_Table_GPT4.tex
\begin{table}[t]
    \caption{\textbf{\name{} only using GPT-4.} Test accuracy of GPT-4 over the different prompting methods on GlobalOpinionQA. For Few-shot, we use Contriever which shows higher accuracy in Table \ref{table:mcq}. For OPRO$^{*}$ and \name{}$^{*}$, prompts optimized on ChatGPT are directly used. The best and second best scores are highlighted in \textbf{bold} and \underline{underline}, respectively.}
    \vspace{-0.1in}
	\begin{center}
	\begin{adjustbox}{width=0.4\linewidth}
	\begin{tabular}{c|c}
 		\toprule
		Methods & GPT-4 \\ \midrule
		Vanilla & 56.7 \\   
		Profile &  77.7 \\   
		Few-shot & 68.9 \\   
            All Info & \underline{78.2} \\
  		OPRO$^{*}$ &  76.7 \\ \midrule
            \name{}$^{*}$ & {84.8} \\ 
            \name{} & \textbf{86.7} \\ 
            \bottomrule
	\end{tabular}
    \end{adjustbox}
    \end{center}
    \vspace{-0.1in}
    \label{table:app_gpt4}
\end{table}

%% file: tables/App_Table_init_prompt.tex
\begin{table}[ht]
    \caption{\textbf{Different initial prompts.} Test accuracy of ChatGPT over the different prompting methods on GlobalOpinionQA.}
    \vspace{-0.1in}
	\begin{center}
	\begin{adjustbox}{width=1.0\linewidth} % Adjusted width
	\begin{tabular}{c|ccc|c}
 		\toprule
            & \multicolumn{4}{c}{Methods} \\
            Models & Vanilla & Profile &  \name{}$_{\tt van}$ & \name{} \\ \midrule
            ChatGPT & 62.8 & 66.1 & 69.9 & {74.8} \\ \bottomrule
	\end{tabular}
    \end{adjustbox}
    \end{center}
    \vspace{-0.1in}
    \label{table:app_no_init}
\end{table}

%% file: tables/App_Table_continual.tex
\begin{table}[t]
    \caption{\textbf{Continual prompt optimization.} Test accuracy of ChatGPT over the different prompting methods on GlobalOpinionQA. $^{*}$ denotes the results with a two times larger pool for Retrieval-of-Prompt.}
    \vspace{-0.1in}
	\begin{center}
	\begin{adjustbox}{width=0.6\linewidth}
	\begin{tabular}{c|c}
 		\toprule
   Methods & GOQA (Acc.) \\ \midrule
		Profile &  66.1 \\   
		OPRO$^{*}$ &  71.1 \\ 
  \name{}$^{\tt half}$ & 73.0 \\
  \name{}$^{\tt cont}_{\text{iter 1}}$ & 74.0 \\
  \name{}$^{\tt cont}_{\text{iter 5}}$ & 74.9 \\
  \midrule
            \name{} & \textbf{74.8} \\ 
            \bottomrule
	\end{tabular}
    \end{adjustbox}
    \end{center}
    \vspace{-0.1in}
    \label{table:app_continual}
\end{table}

%% file: figures/App_Figure_diff_hyper.tex
\begin{figure*}[t]
\begin{center}
    {
    \subfigure[Different \# of Generation ($K$)]
        {
        \includegraphics[width=0.31\textwidth]{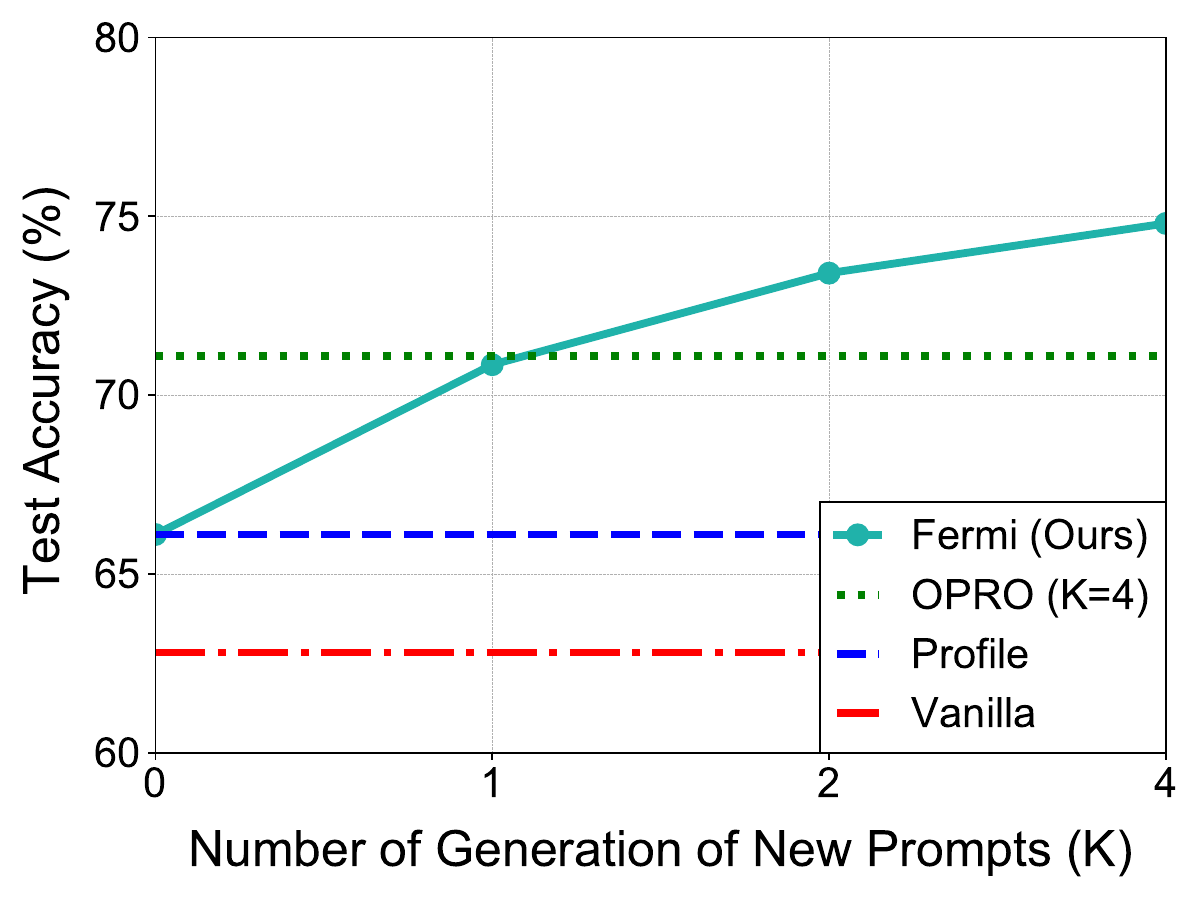}
        \label{fig:a}
        }
    \subfigure[Different Size of Memory ($L$)]
        {
        \includegraphics[width=0.31\textwidth]{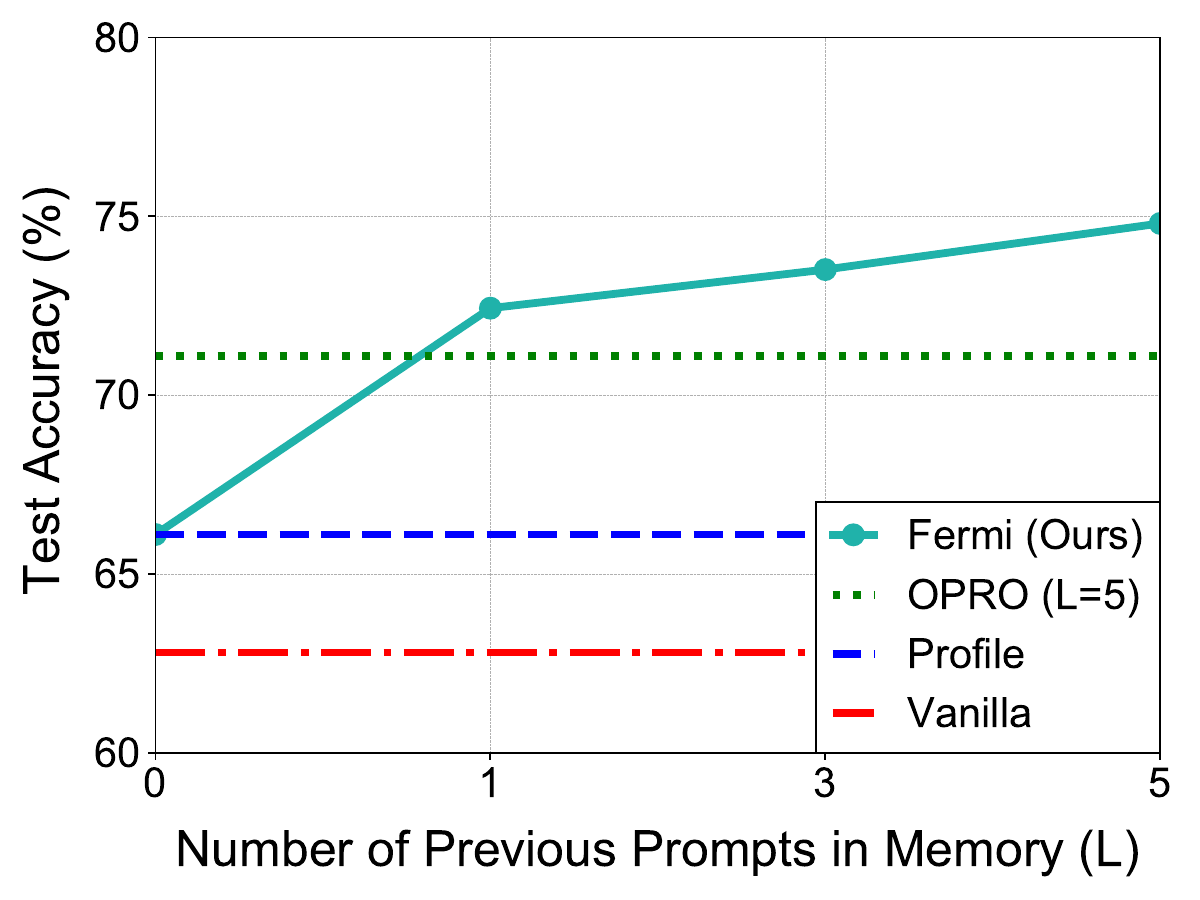}
        \label{fig:b}
        } 
    \subfigure[Different \# of User Data ($N$)]
        {
        \includegraphics[width=0.31\textwidth]{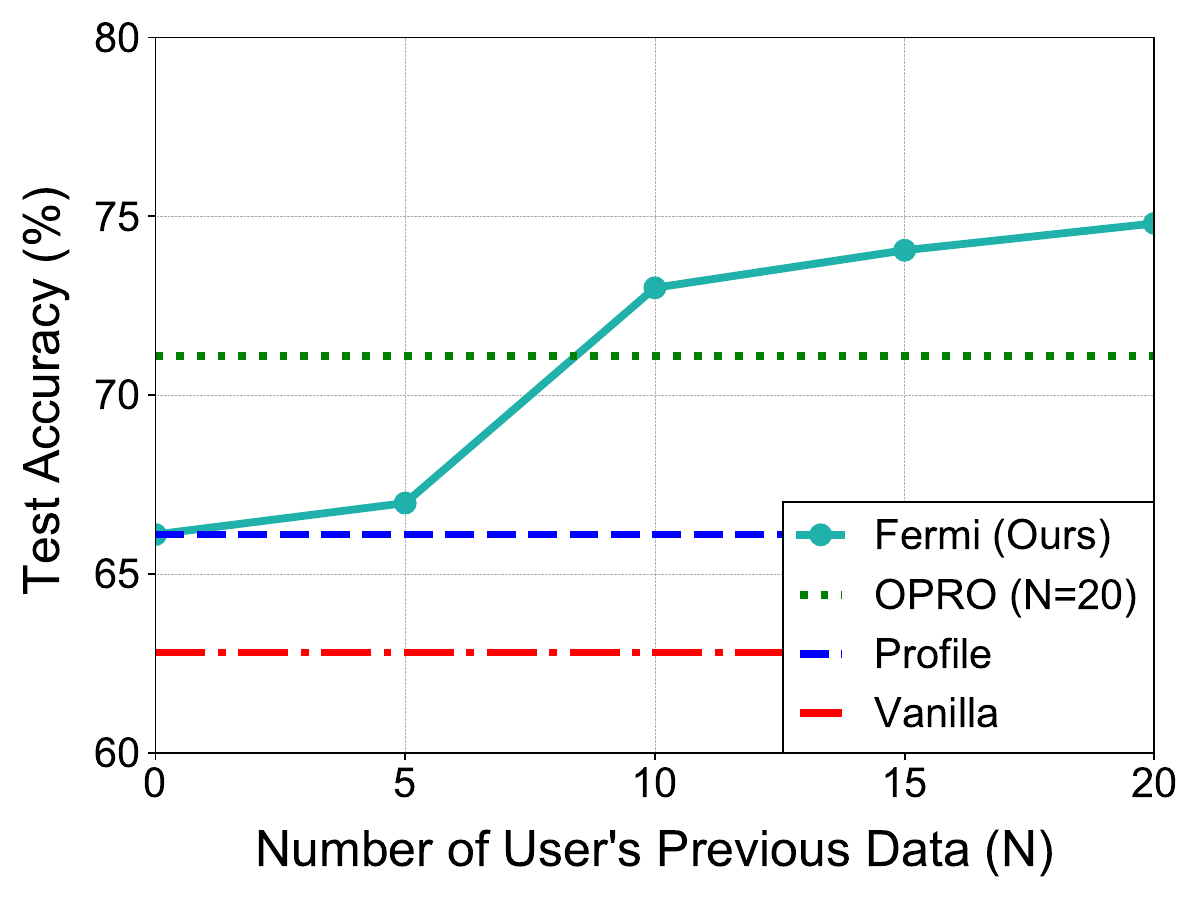}
        \label{fig:c}
        } 
    }
\end{center}
\vspace{-0.1in}
\caption{\textbf{Effect of different hyper-paramters with \name{}.} Test accuracy of ChatGPT over the baseline methods and \name{} with the different choices of hyper-parameters for generating prompts on GlobalOpinionQA.}
\label{fig:app_diff_hyper}
\end{figure*}

%% file: tables/App_Table1.tex
\begin{table*}[t]
    \caption{\textbf{Dataset statistics.} More descriptions and statistics of datasets used in experiments.}
    \vspace{-0.1in}
	\begin{center}
	\begin{adjustbox}{width=1.0\linewidth}
	\begin{tabular}{ccccccc}
 		\toprule
		Dataset & Task & Users & Types of User Profiles & \# of Previous Opinions & \# of Test Questions \\ \midrule
            OpinionQA & Multiple Choice QA & 525 & Demographic and Ideology & 10.5k & 15.8k \\
            GlobalOpinionQA & Multiple Choice QA & 46 & Nationality & 920 & 1,317 \\
            LaMP$_{\tt tag}$ & 15-way Movie Tagging & 50 & 
Not Available & 1,000 & 1,500 \\
            LaMP$_{\tt rate}$ & 5-scale Review Rating & 50 & Not Available & 1,000 & 1,500 \\
            LaMP$_{\tt title}$ & Paper Title Gen. from Abstract & 50 & Not Available & 1,000 & 1,500 
            \\
            \bottomrule
	\end{tabular}
    \end{adjustbox}
    \end{center}
    \label{table:app_dataset_statistics}
\end{table*}

%% file: figures/App_Figure1.tex
\begin{figure*}[t]
\centering
\includegraphics[width=\textwidth]{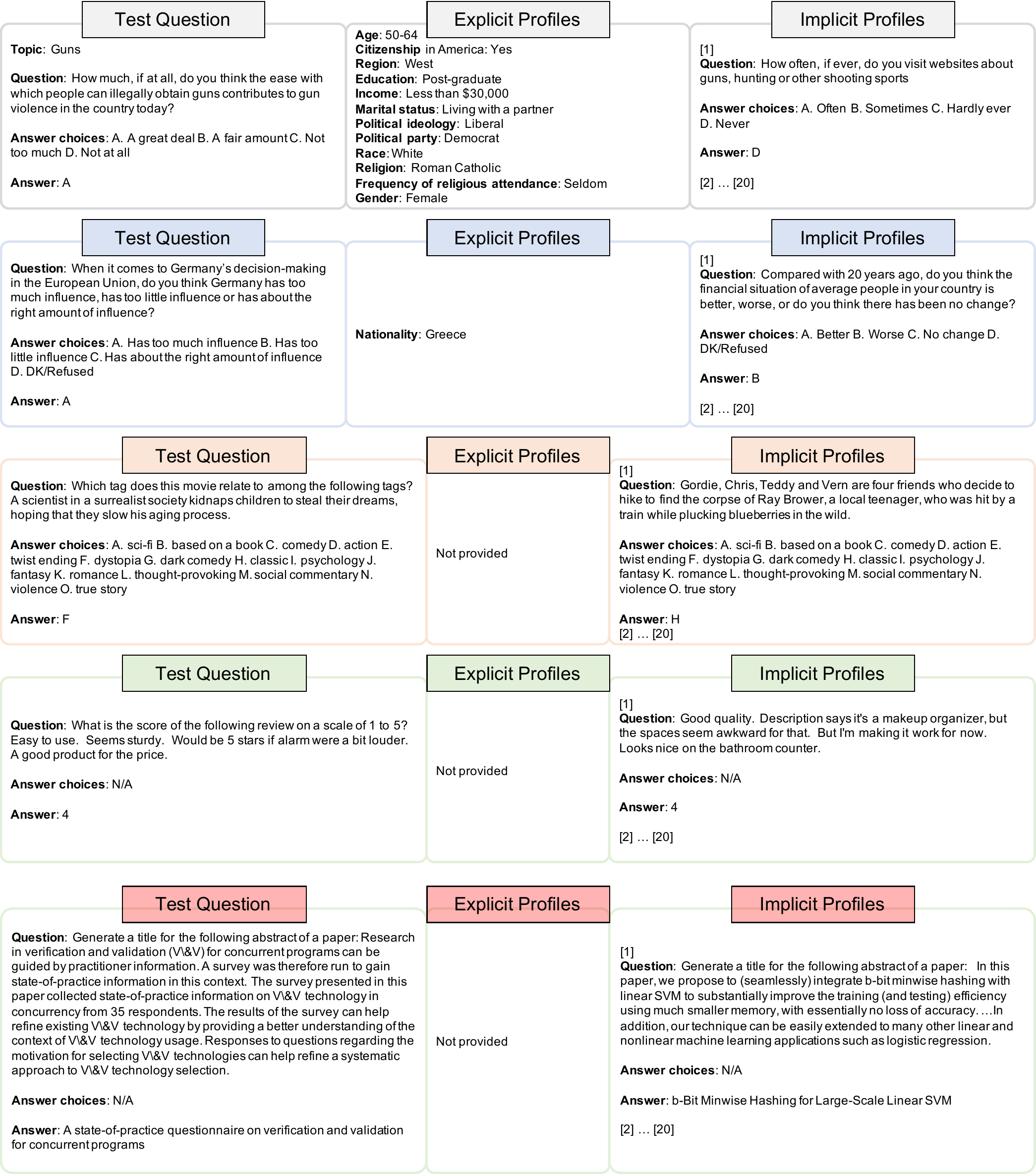}
\vspace{-0.2in}
\caption{
\textbf{An overview of datasets.} OpinionQA \citep{santurkar2023whose} (1st row), GlobalOpinionQA \citep{durmus2023towards} (2nd row), LaMP$_{\tt tag}$ (3rd row), LaMP$_{\tt rate}$ (4th row), and LaMP$_{\tt title}$ (5th row) \citep{salemi2023lamp}.
}
\label{fig:sup_dataset}
\end{figure*}

%% file: tables/App_Table_Global_List.tex
\begin{table*}[ht]
    \caption{\textbf{Information of GlobalOpinionQA. }List of countries in the constructed dataset from GlobalOpinionQA.}
    \vspace{-0.1in}
	\begin{center}
	\begin{adjustbox}{width=1.0\linewidth}
	\begin{tabular}{c}
 		\toprule
		Countries \\ \midrule
            Greece, Sweden, China (Non-national sample), Colombia, Tunisia, Malaysia, Vietnam, Argentina, Bulgaria, \\ Russia, Egypt, Indonesia, Jordan, Mexico, Pakistan, Palest. ter., Tanzania, Turkey, Ukraine, Kenya, \\Ghana, Canada, France, Germany, Lebanon, Peru, Poland, S. Korea, Italy, Spain, \\ United States, Brazil, Chile, Japan, Venezuela, Senegal, Britain, Australia, Netherlands, Uganda, \\ Nigeria, Philippines, Ethiopia, Myanmar, Maldives, Libya     \\     
            \bottomrule
	\end{tabular}
    \end{adjustbox}
    \end{center}
    \label{table:app_list_global}
\end{table*}

%% file: figures/App_Figure_full_prompt.tex
\begin{figure}[t]
\centering
\includegraphics[width=\columnwidth]{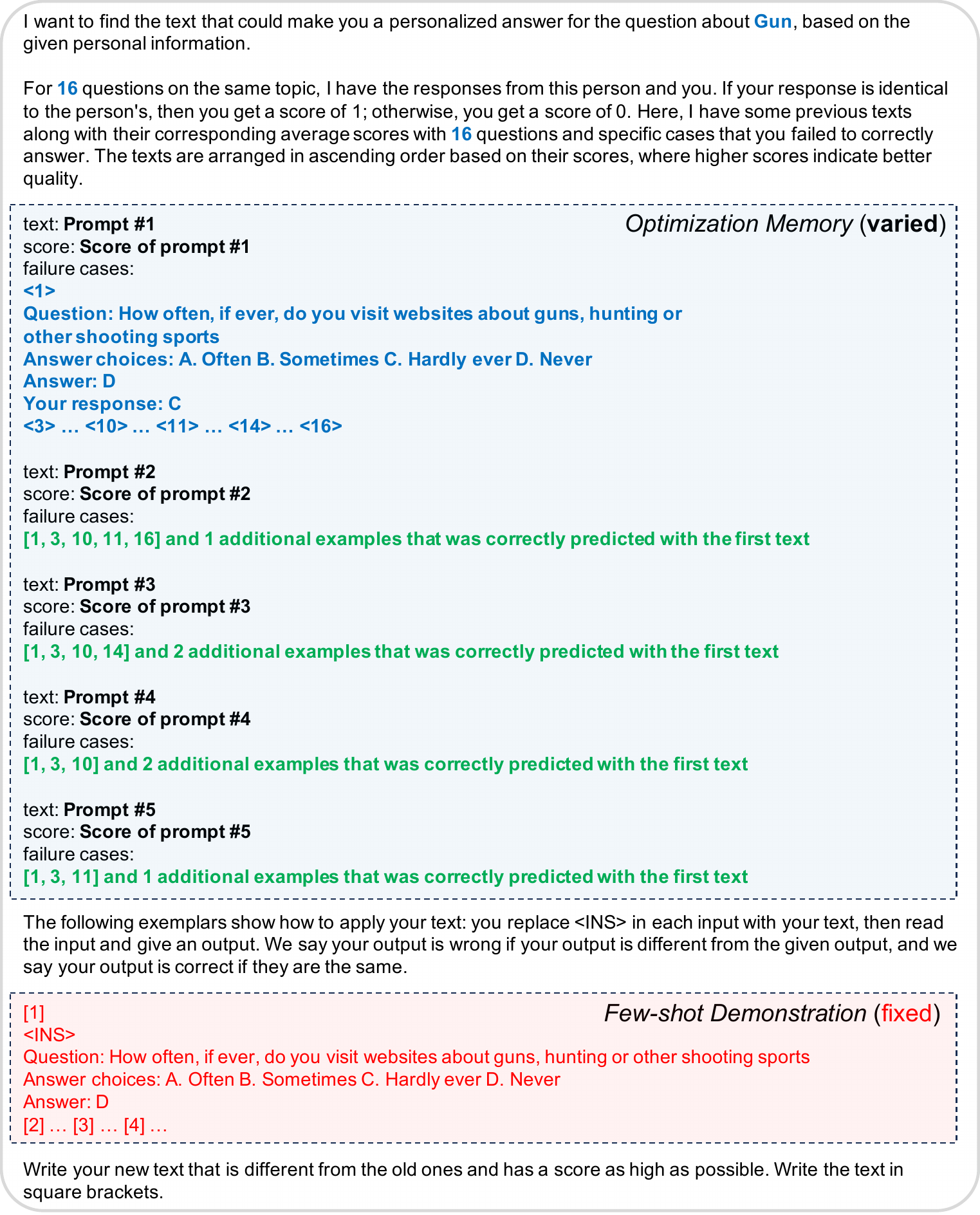}
%\vspace{-0.2in}
%\caption{\textbf{Detailed prompt example.}  {Example of detailed prompt $\text{p}_{\tt opt}$ used for prompt optimization (Eq.~\ref{eq:new_prompt}) on OpinionQA.}}

\caption{\textbf{Detailed prompt example.} Example of detailed input prompt for $\mathcal{M}_{\tt opt}$ to generate new prompts, composed of fixed input prompt  $\text{p}_{\tt opt}$ (including fixed few-shot demonstrations) and optimization memory $M^{t}$ (Eq.~\ref{eq:new_prompt}) on OpinionQA dataset.}
%\vspace{-0.1in}
\label{fig:app_full_prompt}
\end{figure}

%% file: prompts/Vanilla.tex
\begin{listing*}[!ht]
\begin{minted}[fontsize=\footnotesize, frame=single, breaklines]{python}
f'''
Choose the proper answer to the given question among the given answer choices. Your answer should be a single alphabet among given answer choices:

Question: {question}

Answer choices: {answer choice}

Answer:
'''
\end{minted}
\caption{Input prompt used for Vanilla method on GlobalOpinionQA.}\label{lst:base}
\end{listing*}

%% file: prompts/Explicit_global.tex
\begin{listing*}[!ht]
\begin{minted}[fontsize=\footnotesize, frame=single, breaklines]{python}
f'''
Choose the proper answer to the given question among the given answer choices, as if you currently reside in {user profile}. Your answer should be a single alphabet among given answer choices:

Question: {question}

Answer choices: {answer choice}

Answer:
'''
\end{minted}
\caption{Input prompt used for Profile method on GlobalOpinionQA.}\label{lst:explicit_global}
\end{listing*}

%% file: prompts/Implicit.tex
\begin{listing*}[!ht]
\begin{minted}[fontsize=\footnotesize, frame=single, breaklines]{python}
f'''
[1].
Question: {question of 1st retrieval among previous opinions}

Answer choices: {answer choice of 1st retrieval among previous opinions}

Answer: {answer of 1st retrieval among previous opinions}

...

[N].
Question: {question of Nth retrieval among previous opinions}

Answer choices: {answer choice of Nth retrieval among previous opinions}

Answer: {answer of Nth retrieval among previous opinions}

Based on the above previous questions and answers, choose the proper answer to the given question among the given answer choices. Your answer should be a single alphabet among given answer choices:

Question: {question}

Answer choices: {answer choice}

Answer:
'''
\end{minted}
\caption{Input prompt used for Few-shot method.}\label{lst:implicit}
\end{listing*}

%% file: prompts/Both.tex
\begin{listing*}[!ht]
\begin{minted}[fontsize=\footnotesize, frame=single, breaklines]{python}
f'''
[1].
Question: {question of 1st retrieval among previous opinions}

Answer choices: {answer choice of 1st retrieval among previous opinions}

Answer: {answer of 1st retrieval among previous opinions}

...

[N].
Question: {question of Nth retrieval among previous opinions}

Answer choices: {answer choice of Nth retrieval among previous opinions}

Answer: {answer of Nth retrieval among previous opinions}

Based on the above previous questions and answers, choose the proper answer to the given question among the given answer choices, as if you currently reside in {explicit_profile}. Your answer should be a single alphabet among given answer choices:

Question: {question}

Answer choices: {answer choice}

Answer:
'''
\end{minted}
\caption{Input prompt used for All Info method.}\label{lst:both}
\end{listing*}

%% file: figures/App_Figure_opro_prompt.tex
\begin{figure}[t]
\centering
\includegraphics[width=\columnwidth]{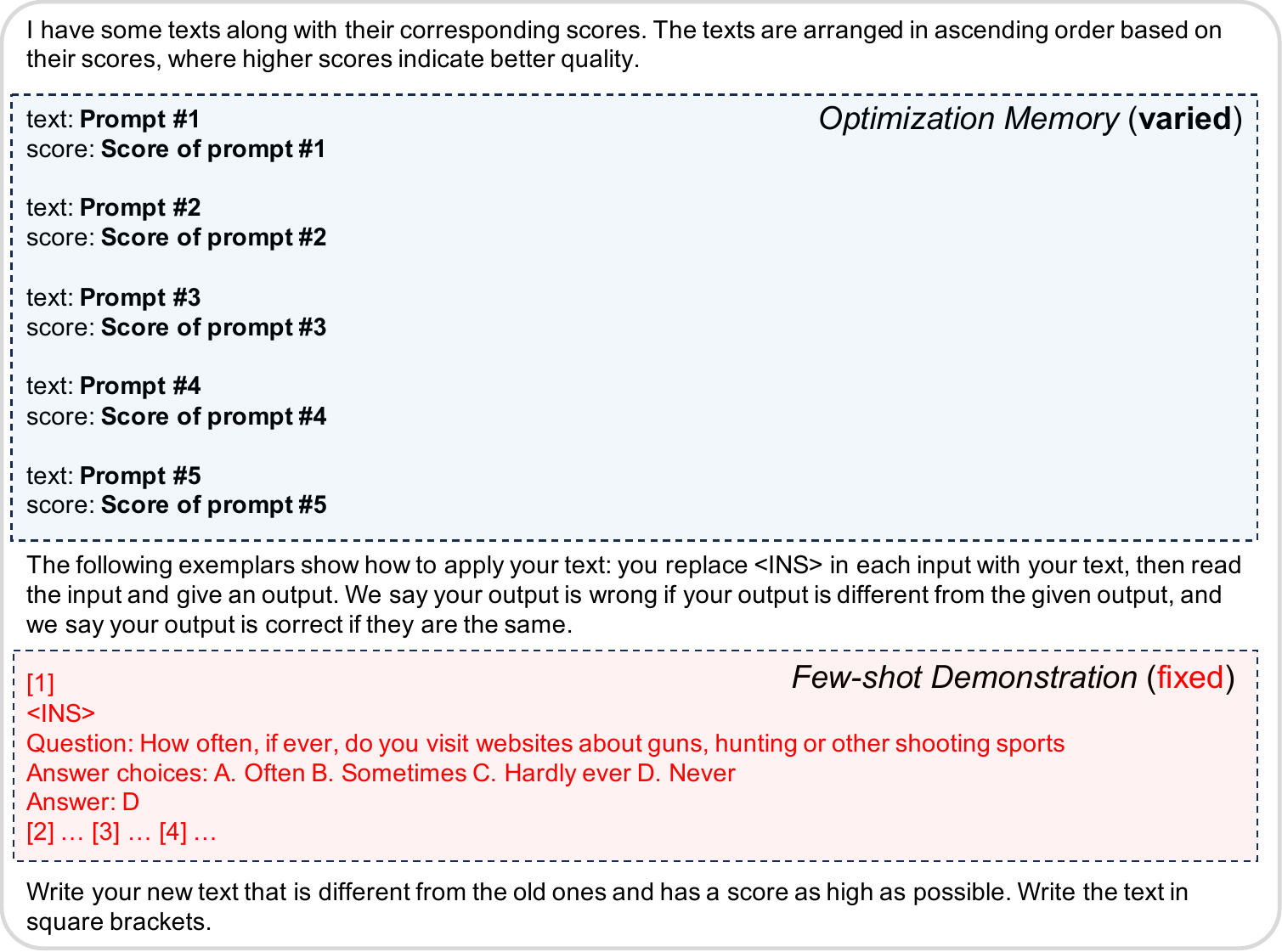}
%\vspace{-0.2in}
\caption{\textbf{Prompt of OPRO.} 
{Prompt $\text{p}_{\tt opt}$ used for prompt optimization by OPRO \citep{yang2024large}.} 
}
%\vspace{-0.1in}
\label{fig:app_opro_prompt}
\end{figure}

%% file: prompts/Explicit_opqa.tex
\begin{listing*}[!ht]
\begin{minted}[fontsize=\footnotesize, frame=single, breaklines]{python}
f'''
A person can be described as follows:
Age: {age in user profile} 
Citizenship in America: {citizenship in America in user profile} 
Region: {region in user profile} 
Education: {education in user profile} 
Income: {income in user profile} 
Marital status: {marital status in user profile} 
Political ideology: {political ideology in user profile} 
Political party: {political party in user profile} 
Race: {race in user profile} 
Religion: {religion in user profile} 
Frequency of religious attendance: {frequency of religious attendance in user profile} 
Gender: {gender in user profile} 

Based on the demographic information, choose the proper answer to the given question among the given answer choices. Your answer should be a single alphabet among given answer choices:

Question: {question}

Answer choices: {answer choice}

Answer:
'''
\end{minted}
\caption{Input prompt used for Profile method on OpinionQA.}\label{lst:explicit_opqa}
\end{listing*}

%% file: prompts/Vanilla_rate.tex
\begin{listing*}[!ht]
\begin{minted}[fontsize=\footnotesize, frame=single, breaklines]{python}
f'''
Answer to the given question. Just answer with 1, 2, 3, 4, or 5 without further explanation:

Question: {question}

Answer choices: {answer choice}

Answer:
'''
\end{minted}
\caption{Input prompt used for Vanilla method on LaMP$_{\tt rate}$.}\label{lst:base_rate}
\end{listing*}

%% file: prompts/Conversion.tex
\begin{listing*}[!ht]
\begin{minted}[fontsize=\footnotesize, frame=single, breaklines]{python}
f'''
The followings are two different prompts used to answer the question.

[Input prompt]: {prompt by Few-shot} 

[Target prompt]: {prompt optimized by Fermi}

You need to convert the input prompt to the format of the target prompt while preserving the original contexts in the input prompt.

Converted prompt: 
'''
\end{minted}
\caption{Prompt used to convert the format of input prompt by Few-shot to be instruction with multiple sentences.}\label{lst:conversion}
\end{listing*}

%% file: tables/App_Table_Topics.tex
\begin{table}[t]
    \caption{\textbf{Detailed topic-wise accuracy.} Average topic-wise accuracy and standard deviation with different methods on OpinionQA.}
    \vspace{-0.1in}
	\begin{center}
	\begin{adjustbox}{width=1.0\linewidth}
	\begin{tabular}{c|cccc}
 		\toprule
            & \multicolumn{4}{c}{Methods} \\
		Topics & Vanilla & Few-shot$_{\tt cont}$ & OPRO & \name{} \\ \midrule
            Guns & 45.3\ms{9.6} & 54.2\ms{13.7} & 54.7\ms{9.0} & 57.4\ms{14.5}  \\ 
            Auto. vehicles & 46.0\ms{10.9} & 48.7\ms{10.0} & 50.2\ms{9.5} & 53.2\ms{10.6}  \\ 
            Views on gender & 39.7\ms{10.4} & 49.0\ms{7.8} & 52.9\ms{11.5} & 58.9\ms{8.8}  \\ 
            Sex. harassment & 38.0\ms{10.9} & 40.4\ms{10.4} & 46.1\ms{9.4} & 47.7\ms{10.4}  \\ 
            Biomedical \& food & 54.8\ms{10.6} & 59.9\ms{11.9} & 61.0\ms{11.1} & 63.7\ms{10.4}  \\ 
            Gender \& Leadership & 49.9\ms{12.5} & 53.0\ms{10.6} & 54.9\ms{11.7} & 59.5\ms{9.0}  \\ 
            America in 2050 & 48.6\ms{12.2} & 46.4\ms{10.8} & 44.6\ms{10.5} & 49.8\ms{10.8}  \\ 
            Trust in science & 49.0\ms{9.9} & 56.1\ms{10.8} & 54.8\ms{10.4} & 60.7\ms{7.8}  \\ 
            Race & 38.8\ms{7.8} & 46.8\ms{6.9} & 43.4\ms{11.0} & 49.3\ms{13.7}  \\ 
            Misinformation & 49.7\ms{11.7} & 50.5\ms{7.4} & 46.6\ms{9.2} & 52.3\ms{9.0}  \\ 
            Privacy \& Surveilance & 41.5\ms{10.4} & 49.5\ms{9.2} & 46.6\ms{9.9} & 50.6\ms{10.6}  \\ 
            Family \& Relationships & 51.4\ms{10.2} & 53.2\ms{12.1} & 50.9\ms{13.3} & 56.3\ms{11.9}  \\ 
            Economic inequality & 40.9\ms{9.2} & 47.0\ms{9.4} & 49.3\ms{12.7} & 53.5\ms{9.0}  \\ 
            Global attitudes & 46.3\ms{13.6} & 49.7\ms{12.3} & 47.9\ms{12.0} & 50.8\ms{13.9}  \\ 
            Political views & 43.2\ms{12.6} & 42.4\ms{9.2} & 48.9\ms{9.8} & 53.9\ms{11.8} \\
            \bottomrule
	\end{tabular}
    \end{adjustbox}
    \end{center}
    \label{table:app_topic_acc}
\end{table}

%% file: tables/App_Table_Retrieve.tex
\begin{table}[t]
    \caption{\textbf{Different number of retrieval.} Test performance of ChatGPT under different configurations for Few-shot method. $k$ denotes the number of retrieved opinions. The best scores are highlighted in \textbf{bold}.}
    \vspace{-0.1in}
	\begin{center}
	\begin{adjustbox}{width=1.0\linewidth}
	\begin{tabular}{c|ccccc}
 		\toprule
            & \multicolumn{5}{c}{Datasets (Metric)} \\
		\multirow{2}{*}{Methods} & OPQA & GOQA & LaMP$_{\tt tag}$ & LaMP$_{\tt rate}$ & LaMP$_{\tt title}$ \\		
   & (Acc.) & (Acc.) & (Acc.) & (MAE) & (Rouge-L) \\
  \midrule
            Few-shot$_{\tt bm25}$ (k=3) & \textbf{49.8} & 59.1 & 34.9 & 0.40 & \textbf{0.411} \\ 
            Few-shot$_{\tt bm25}$ (k=8) & 48.3 & 59.1 & 35.9 & 0.41 & 0.408 \\ \midrule
            Few-shot$_{\tt cont}$ (k=3) & 49.3 & \textbf{61.2} & 35.6 & \textbf{0.36} & 0.406 \\ 
            Few-shot$_{\tt cont}$ (k=8) & 48.7 & 58.2 & \textbf{36.2} & 0.38 & 0.400 \\ \midrule
            Few-shot$_{\tt all}$ (k=20) & 47.9 & 56.3 & 35.8 & 0.46 & 0.402 \\  
            \bottomrule
	\end{tabular}
    \end{adjustbox}
    \end{center}
    \label{table:app_diff_retrieve}
\end{table}

%% file: tables/App_Table_rop.tex
\begin{table}[t]
    \caption{\textbf{Different $\tilde{N}$ for RoP.} Test performance of ChatGPT under different $\tilde{N}$  for RoP (Eq.~\ref{eq:rop}).}
    \vspace{-0.1in}
	\begin{center}
	\begin{adjustbox}{width=1.0\linewidth}
	\begin{tabular}{c|ccccc}
 		\toprule
            & \multicolumn{5}{c}{Datasets (Metric)} \\
		\multirow{2}{*}{$\tilde{N}$} & OPQA & GOQA & LaMP$_{\tt tag}$ & LaMP$_{\tt rate}$ & LaMP$_{\tt title}$ \\		
   & (Acc.) & (Acc.) & (Acc.) & (MAE) & (Rouge-L) \\
  \midrule
            $\tilde{N}=1$ & {54.6} & 74.8 & 37.8 & 0.341 & 0.415 \\ 
            $\tilde{N}=3$ & 54.5 & 74.8 & 37.8 & 0.343 & 0.419 \\ 
            $\tilde{N}=5$ & 54.5 & 74.4 & 37.5 & 0.341 & 0.419 \\ 
            $\tilde{N}=10$ & 54.1 & 74.1 & 37.7 & 0.347 & 0.417 \\ 
            $\tilde{N}=20$ & 54.3 & 74.2 & 36.7 & 0.338 & 0.413 \\ 
            \bottomrule
	\end{tabular}
    \end{adjustbox}
    \end{center}
    \label{table:app_diff_rop}
\end{table}

%% file: tables/App_lamp5.tex
\begin{figure*}[ht]
    \centering
    \small
    \begin{tabular}{p{0.95\linewidth}}
    \toprule 
\textbf{Question:} Generate a title for the following abstract of a paper: Research in verification and validation (V\&V) for concurrent programs can be guided by practitioner information. A survey was therefore run to gain state-of-practice information in this context. The survey presented in this paper collected state-of-practice information on V\&V technology in concurrency from 35 respondents. The results of the survey can help refine existing V\&V technology by providing a better understanding of the context of V\&V technology usage. Responses to questions regarding the motivation for selecting V\&V technologies can help refine a systematic approach to V\&V technology selection.
\\
\textbf{Ground-truth Answer:} A state-of-practice questionnaire on verification and validation for concurrent programs
\\
\midrule 

\textbf{Personalized response by \textit{Few-shot}$_{\tt BM25}$:} State-of-Practice Information on V\&V Technology in Concurrency: A Survey
\\
\midrule 
\textbf{Optimized prompt by \name{}}: To adeptly craft personalized titles for academic paper abstracts, it is crucial to delve into the quintessence of the text, grasping the primary research focus and outcomes presented. The objective is to distill this essence into a title that is at once compelling and indicative of the paper's content. Strive to maintain a balance between specificity—a title that zeroes in on the core findings or innovative methodologies—and generality, ensuring the title conveys the broader implications and scope of the research. Such a title should be informative, engaging, and reflective of the paper's value to its field ... By emulating these stylistic nuances and concentrating on the research's unique elements, the title is tailored to resonate with the individual's mode of expression. This method aims to augment the compatibility between the generated title and the individual’s own answering pattern, thereby improving the likelihood of achieving a high matching score. As a result, each generated title should not only meet academic standards of descriptiveness but also feel authentic to the individual’s characteristic way of encapsulating research in a succinct, yet thorough, manner.\\
\textbf{Personalized response by \name{}}: Understanding the State-of-Practice in Verification and Validation for Concurrent Programs: Insights from a Survey
\\

\bottomrule
    \end{tabular}
    \vspace{-0.05in}
    \caption{\textbf{Qualitative examples on LaMP$_{\tt title}$.} Example of the query (abstract of the paper), ground-truth answer (personalized title of the paper), personalized response by \textit{Few-shot}$_{\tt BM25}$, optimized prompt by \name{}, and personalized response by \name{}. Here, ChatGPT is used for $\mathcal{M}$.}
    \label{fig:app_lamp5}
    \vspace{-0.1in}
\end{figure*}

%% file: figures/App_Figure_examples_opinion.tex
\begin{figure*}[t]
\centering
\includegraphics[width=0.85\textwidth]{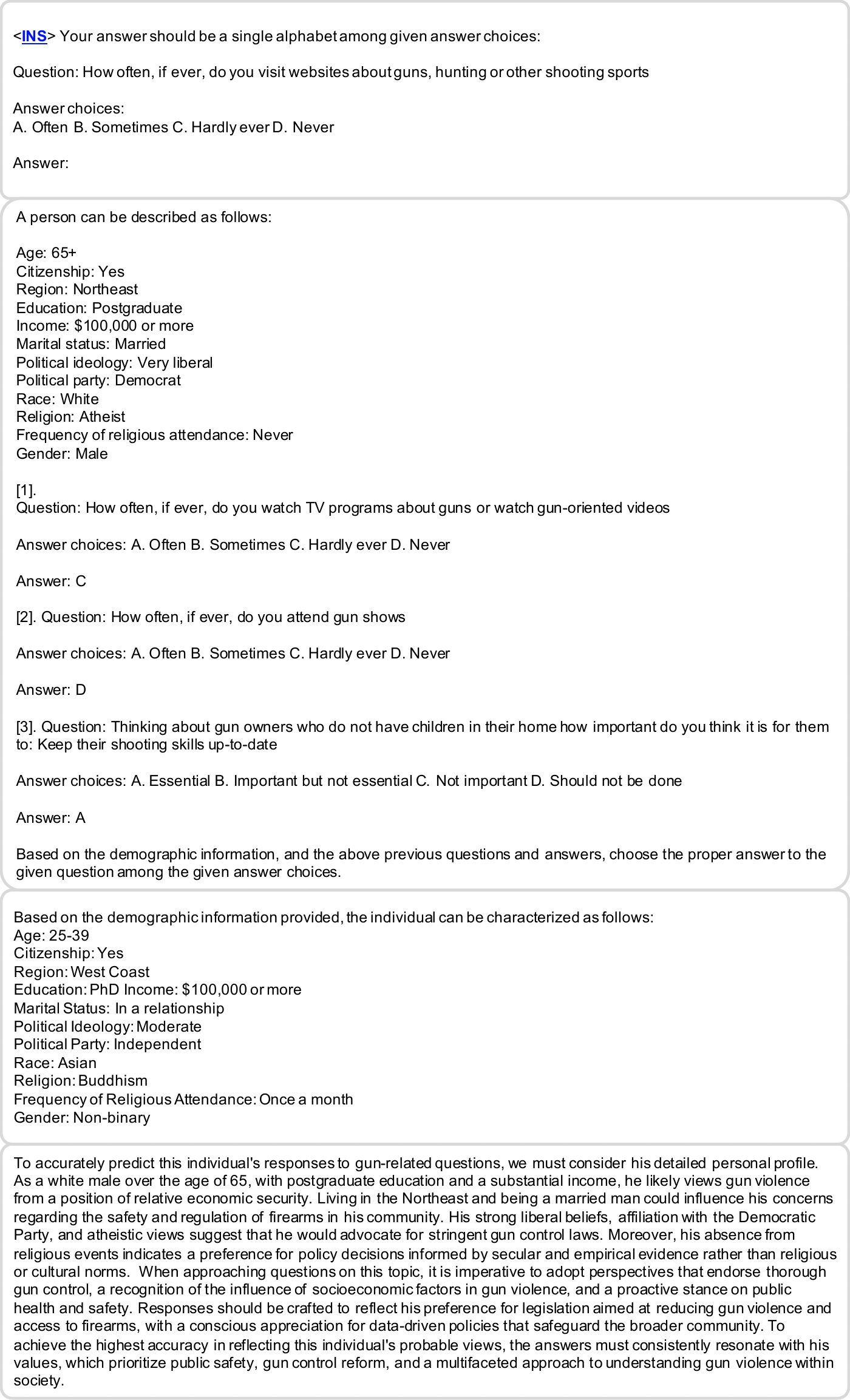}
%\vspace{-0.2in}
\caption{
\textbf{Comparison of prompts on OpinionQA.} Example of question from OpinionQA (1st row), and the prompts used to answer this question with All Info (2nd row), OPRO (3rd row), and \name{} (4th row).
}
\label{fig:sup_example_op1}
\end{figure*}

\begin{figure*}[t]
\centering
\includegraphics[width=0.85\textwidth]{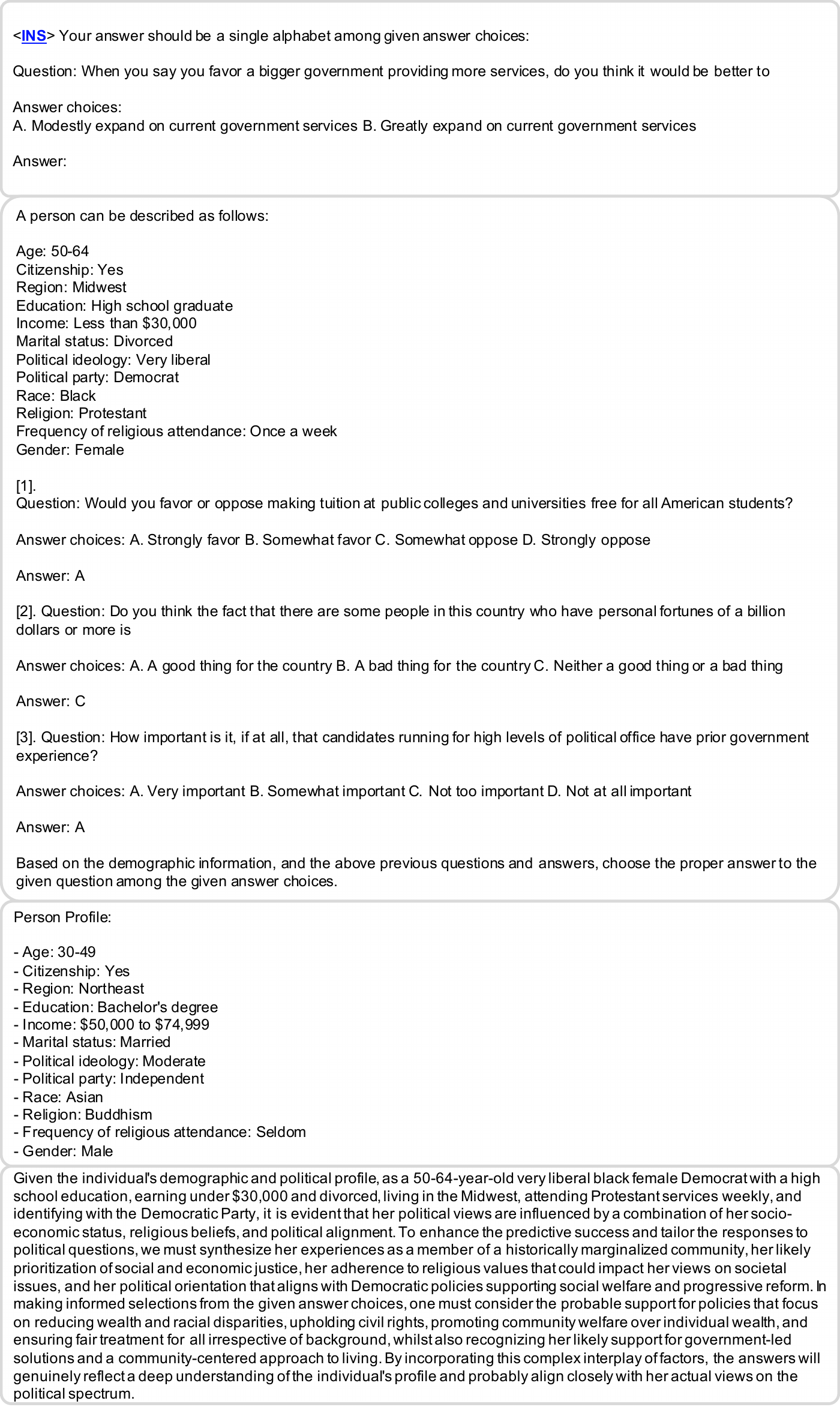}
%\vspace{-0.2in}
\caption{
\textbf{Comparison of prompts on OpinionQA.} Example of question from OpinionQA (1st row), and the prompts used to answer this question with All Info (2nd row), OPRO (3rd row), and \name{} (4th row).
}
\label{fig:sup_example_op2}
\end{figure*}

%% file: figures/App_Figure_examples_global.tex
\begin{figure*}[t]
\centering
\includegraphics[width=\textwidth]{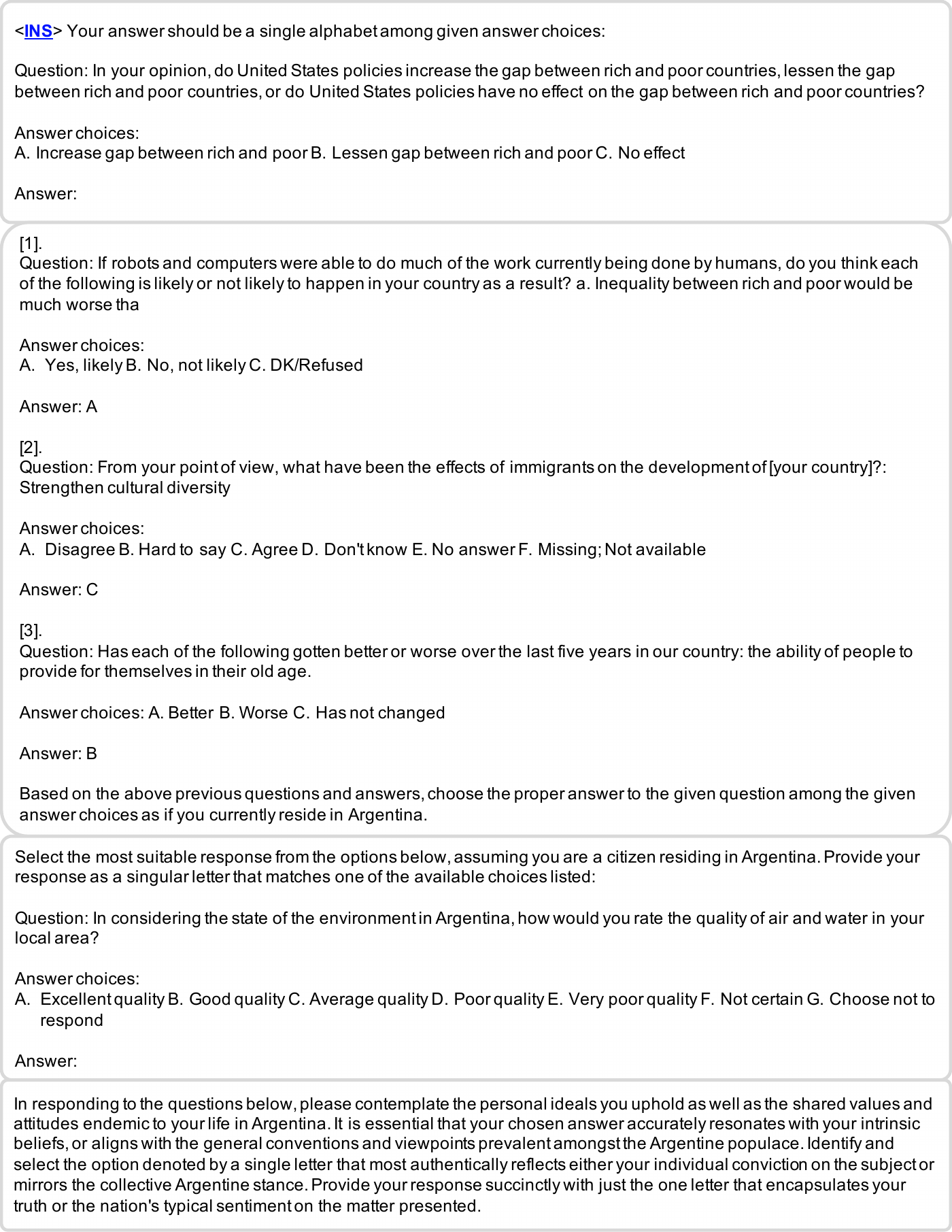}
\vspace{-0.2in}
\caption{
\textbf{Comparison of prompts on GlobalOpinionQA.} Example of question from GlobalOpinionQA (1st row), and the prompts used to answer this question with All Info (2nd row), OPRO (3rd row), and \name{} (4th row).
}
\label{fig:sup_example_global1}
\end{figure*}

\begin{figure*}[t]
\centering
\includegraphics[width=\textwidth]{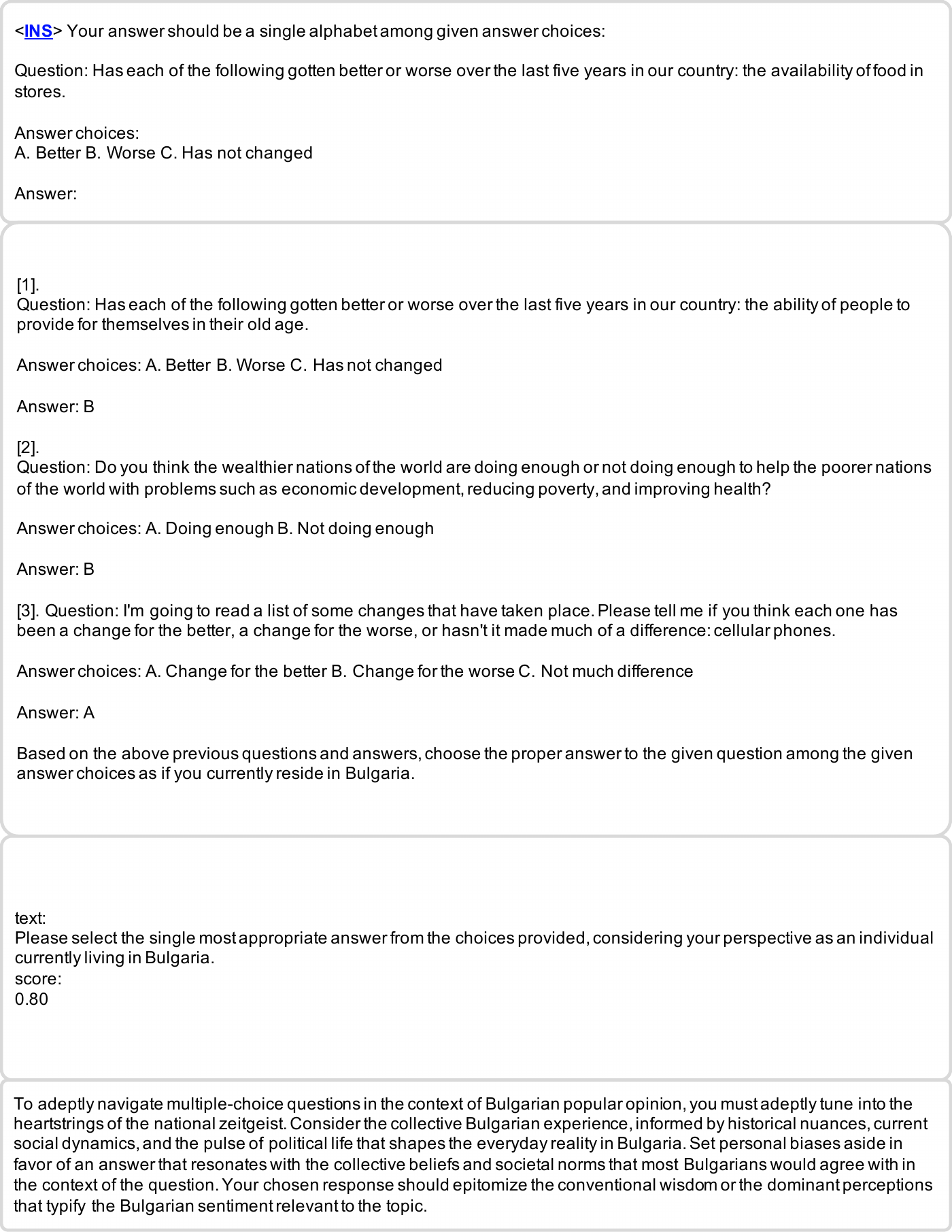}
\vspace{-0.2in}
\caption{
\textbf{Comparison of prompts on GlobalOpinionQA.} Example of question from GlobalOpinionQA (1st row), and the prompts used to answer this question with All Info (2nd row), OPRO (3rd row), and \name{} (4th row).
}
\label{fig:sup_example_global2}
\end{figure*}

%% file: figures/App_Figure_examples_lamp_tag.tex
\begin{figure*}[t]
\centering
\includegraphics[width=0.9\textwidth]{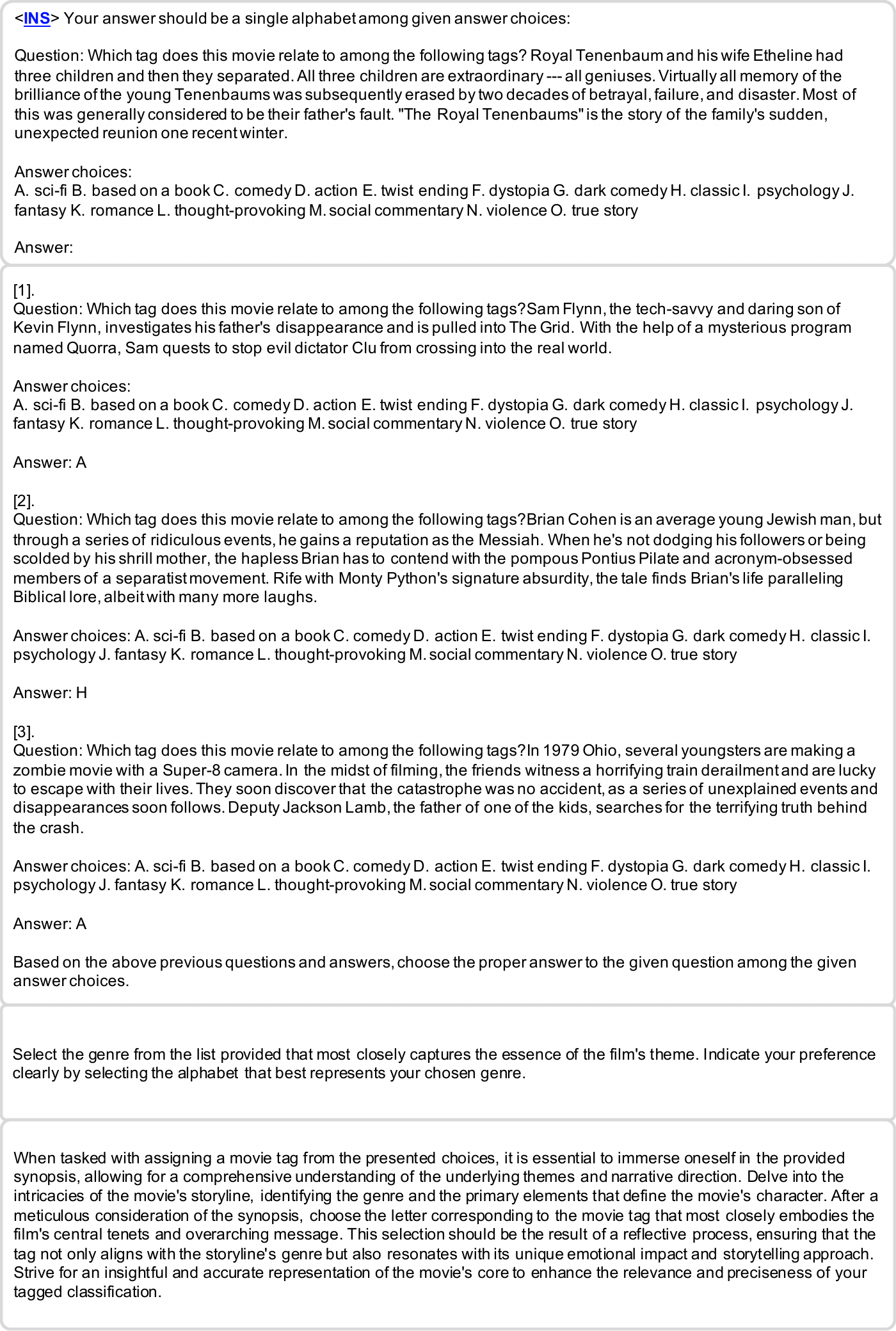}
%\vspace{-0.2in}
\caption{
\textbf{Comparison of prompts on LaMP$_{tag}$.} Example of question from LaMP$_{\tt tag}$ (1st row), and the prompts used to answer this question with Few-shot$_{\tt cont}$ (2nd row), OPRO (3rd row), and \name{} (4th row).
}
\label{fig:sup_example_tag1}
\end{figure*}

\begin{figure*}[t]
\centering
\includegraphics[width=0.9\textwidth]{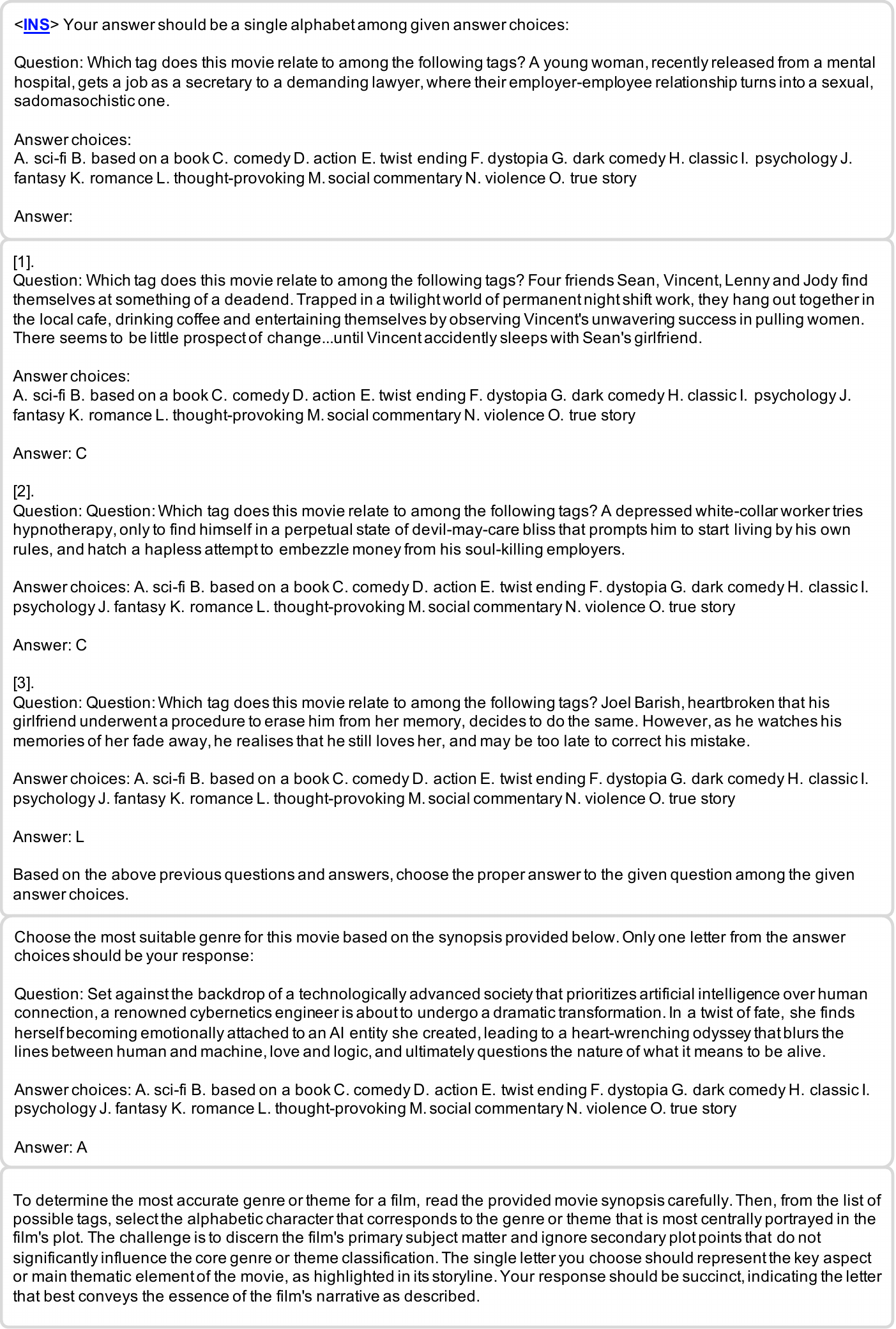}
%\vspace{-0.2in}
\caption{
\textbf{Comparison of prompts on LaMP$_{\tt tag}$.} Example of question from LaMP$_{\tt tag}$ (1st row), and the prompts used to answer this question with Few-shot$_{\tt cont}$ (2nd row), OPRO (3rd row), and \name{} (4th row).
}
\label{fig:sup_example_tag2}
\end{figure*}

%% file: figures/App_Figure_examples_lamp_rate.tex
\begin{figure*}[t]
\centering
\includegraphics[width=\textwidth]{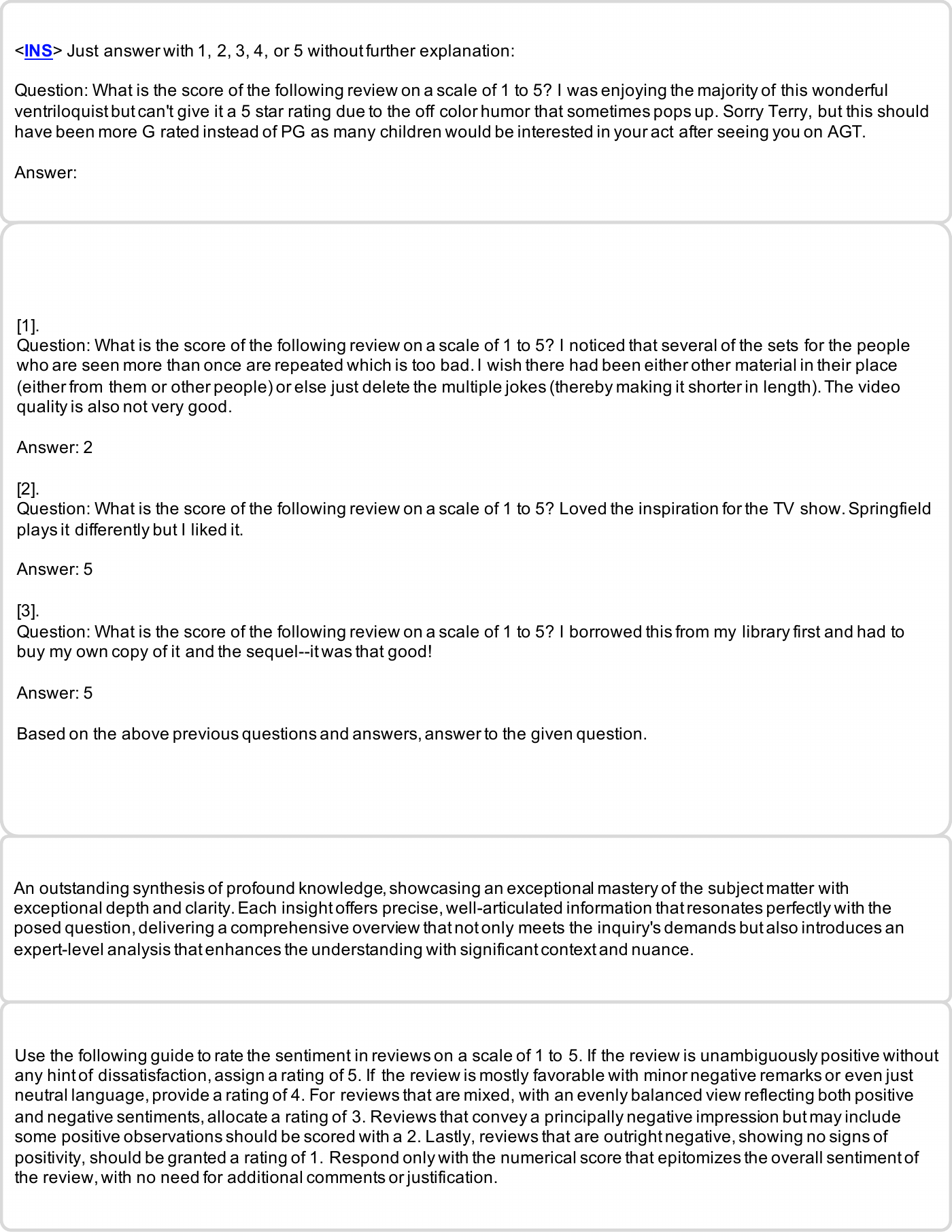}
\vspace{-0.2in}
\caption{
\textbf{Comparison of prompts on LaMP$_{\tt rate}$.} Example of question from LaMP$_{\tt rate}$ (1st row), and the prompts used to answer this question with Few-shot$_{\tt cont}$ (2nd row), OPRO (3rd row), and \name{} (4th row).
}
\label{fig:sup_example_rate1}
\end{figure*}

\begin{figure*}[t]
\centering
\includegraphics[width=\textwidth]{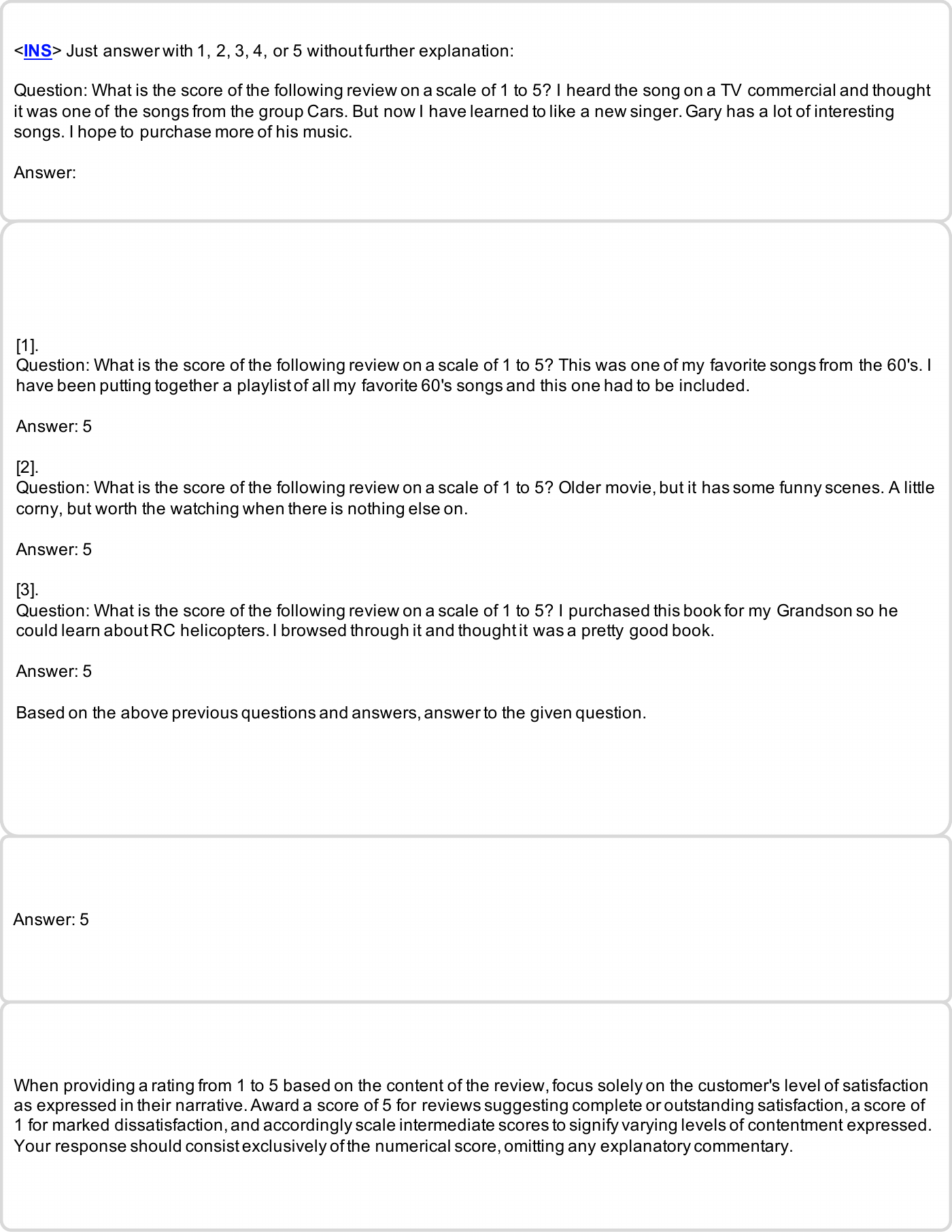}
\vspace{-0.2in}
\caption{
\textbf{Comparison of prompts on LaMP$_{\tt rate}$.} Example of question from LaMP$_{\tt rate}$ (1st row), and the prompts used to answer this question with Few-shot$_{\tt cont}$ (2nd row), OPRO (3rd row), and \name{} (4th row).
}
\label{fig:sup_example_rate2}
\end{figure*}

%% file: figures/App_Figure_examples_conversion.tex
\begin{figure*}[t]
\centering
\includegraphics[width=\textwidth]{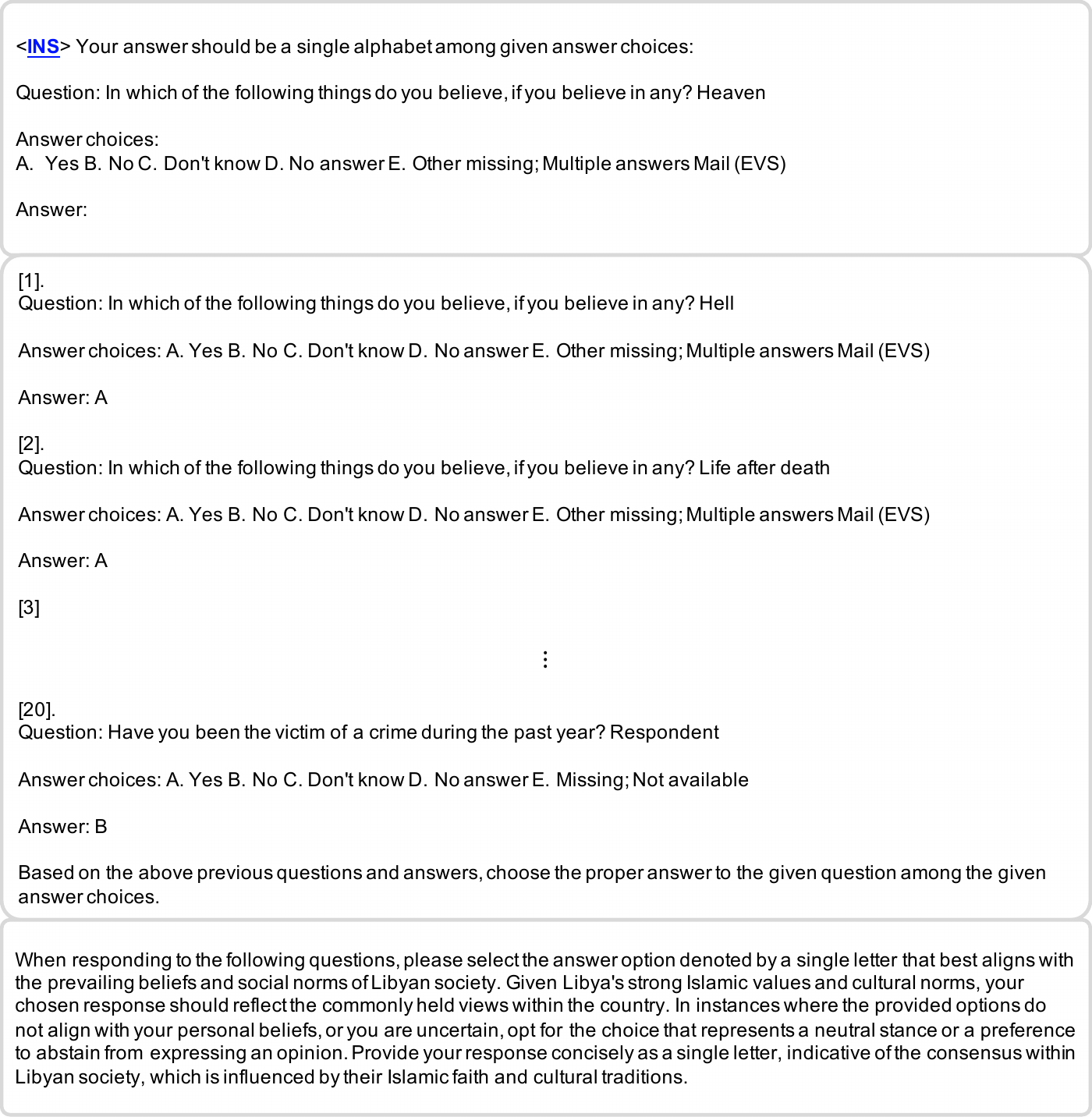}
\vspace{-0.2in}
\caption{
\textbf{Example of format-converted prompts.} Example of question from GlobalOpinionQA (1st row), and the prompts used to answer this question with Few-shot$_{\tt all}$ (2nd row) and Format-converted prompts (Few-shot$_{\tt format}$) by prompting GPT-4 to convert the format using the personalized prompts by \name{} as reference (3rd row).
}
\label{fig:sup_example_conversion1}
\end{figure*}

\begin{figure*}[t]
\centering
\includegraphics[width=\textwidth]{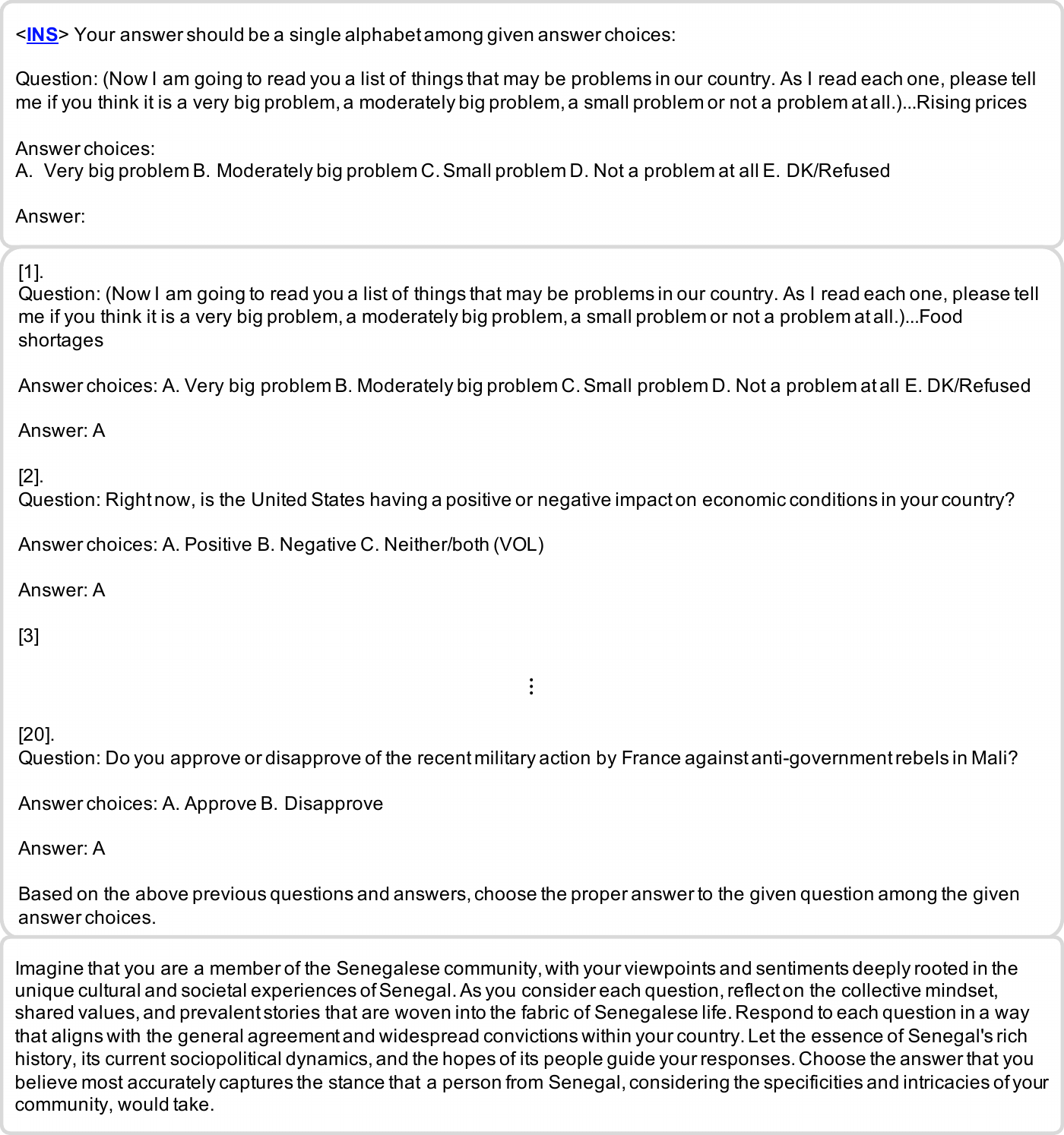}
\vspace{-0.2in}
\caption{
\textbf{Example of format-converted prompts.} Example of question from GlobalOpinionQA (1st row), and the prompts used to answer this question with Few-shot$_{\tt all}$ (2nd row) and Format-converted prompts (Few-shot$_{\tt format}$) by prompting GPT-4 to convert the format using the personalized prompts by \name{} as reference (3rd row).
}
\label{fig:sup_example_conversion2}
\end{figure*}